\documentclass[letterpaper, 10 pt, conference]{ieeeconf}
\pdfoutput=1
\usepackage{graphicx}
\usepackage{epstopdf}
\usepackage[utf8]{inputenc}
\usepackage{pgfplots}
\usepackage{amsmath,amssymb}
\usepackage{subfig}
\usepackage{xfrac}
\usepackage[letterpaper, left=0.667in, right=0.667in, bottom=0.597in, top=0.833in]{geometry}
\usepackage{algorithm}
\usepackage[noend]{algpseudocode}
\usepackage{bm}

\pgfplotsset{compat=1.17}

\usepackage{xcolor}

\title{Online Coverage Planning for an Autonomous Weed Mowing\\ Robot with Curvature Constraints}

\author{Parikshit Maini, Burak M. Gonultas and Volkan Isler\\
  Department of Computer Science\\
  University of Minnesota - Twin Cities, USA\\
  \texttt{\{pmaini,gonul004,isler\}@umn.edu}\\
}

\date{August 2021}

\begin{document}

\maketitle

\begin{abstract}
The land used for grazing cattle takes up about one-third of the land in the United States. These areas can be highly rugged. Yet, they need to be maintained to prevent weeds from taking over the nutritious grassland. This can be a daunting task especially in the case of organic farming since herbicides cannot be used. In this paper, we present the design of Cowbot, an autonomous weed mowing robot for pastures. Cowbot is an electric mower designed to operate in the rugged environments on cow pastures and provide a cost-effective method for weed control in organic farms. 
  
Path planning for the Cowbot is challenging since weed distribution on pastures is unknown. Given a limited field of view, online path planning is necessary to detect weeds and plan paths to mow them. We study the general online path planning problem for an autonomous mower with curvature and field of view constraints.
We develop two online path planning algorithms that are able to utilize new information about weeds to optimize path length and ensure coverage. We deploy our algorithms on the Cowbot and perform field experiments to validate the suitability of our methods for real-time path planning. We also perform extensive simulation experiments which show that our algorithms result in up to 60 \% reduction in path length as compared to baseline boustrophedon and random-search based coverage paths.
    



    
  
    
    
\end{abstract}
\section{Introduction}


Dairy pastures constitute about one third of the land in the United States \cite{ag_census}. A single dairy farm averages around 460 acres. Maintaining these large pastures to prevent weeds from taking over the nutritious grasslands is a daunting task, especially in organic farms that do not use herbicides. Manual mowing is strenuous and unpleasant. Many recent projects have developed robotic weeding solutions for outdoor \cite{bawdenAGBOT,evert2011jfr,PEREZRUIZ201445,HANSEN2013217} and indoor (greenhouse) \cite{OBERTI2016203,BARTH201671} weed control applications in agriculture. However, robotic weed removal efforts to-date have primarily focused on crop-cultivating farms that practice row-plantation. By virtue of row plantations, the robots have a strong prior on the outlay of the field and the location of weeds - inter-row and intra-row. Cow pastures on the other hand are usually large open fields with little to no regular maintenance. In the case of organic dairy farms, that form the focus group for this work, weed growth is significant and can spread fast if not controlled in time. Rough uneven terrain combined with large and dense weed populations makes it challenging to operate in cow pastures. This results in the need to develop a robust and intelligent robotic weeding platform. 


\begin{figure}
    \centering 
    \subfloat{\includegraphics[trim=3cm 7cm 5cm 0cm,clip,height = 2.5cm]{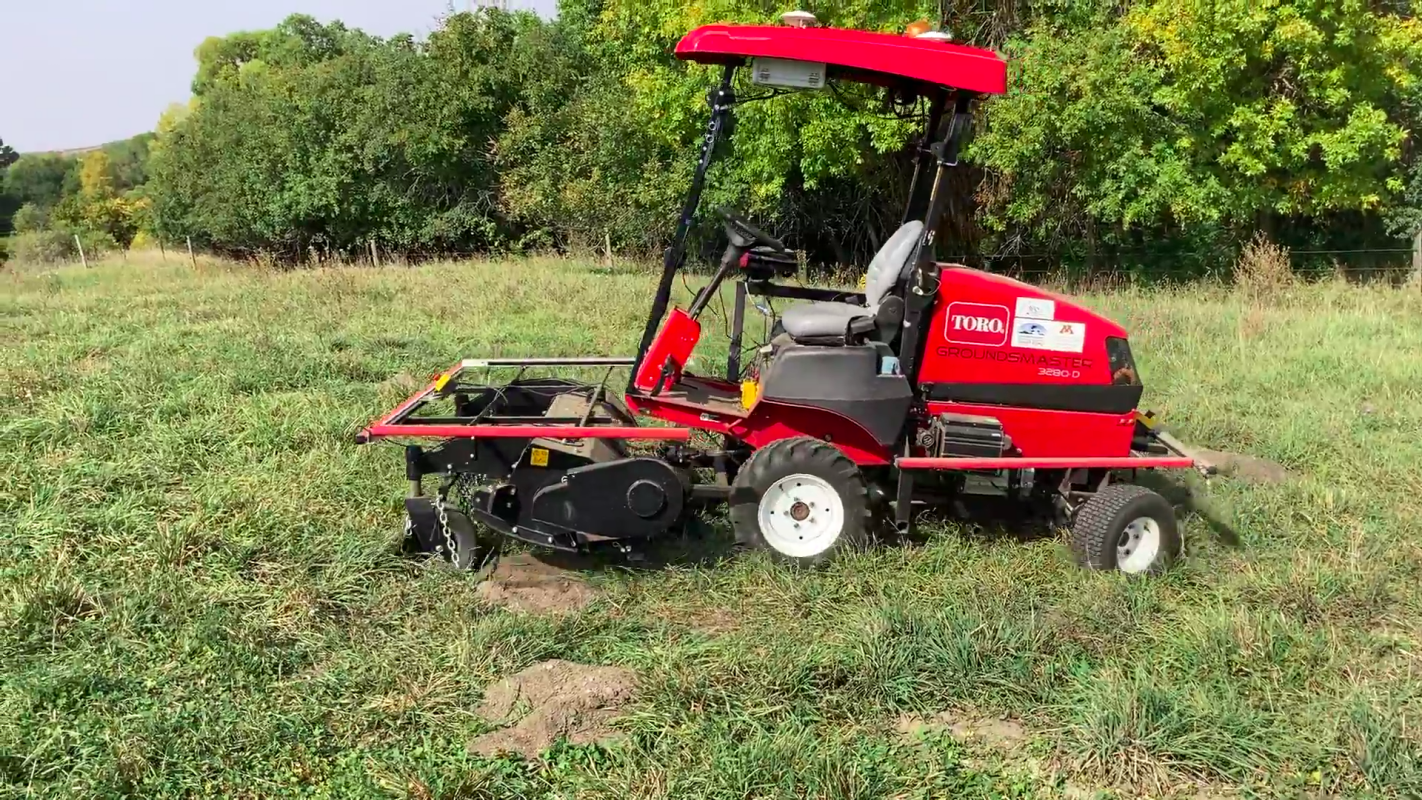}\label{fig:robot_model}}
    \hspace{1mm}
    \subfloat{\includegraphics[height = 2.5cm]{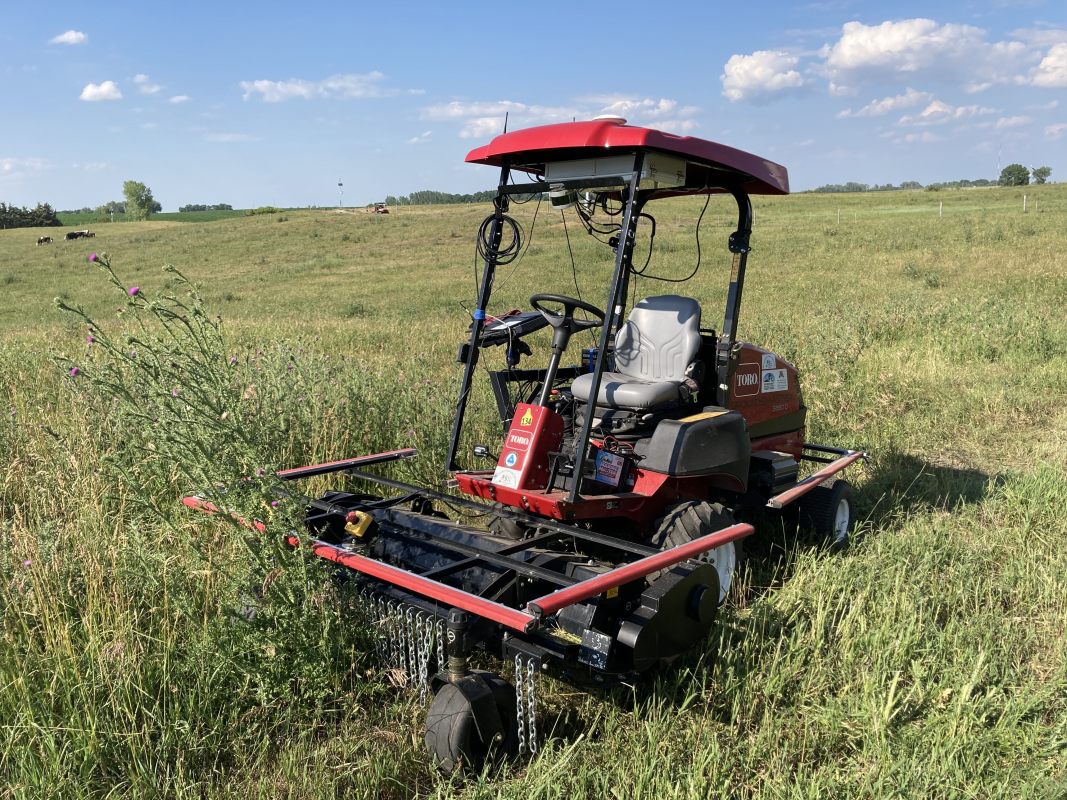}\label{fig:robot_model}}
    \caption{Cowbot - autonomous weed mower operating on a cow pasture. As shown in the image on the right, weeds on cow pastures can grow up to 1-2 meters tall.}\label{fig:cowbot_profile}
\end{figure}



In this paper, we present the design of our autonomous weed mowing robot for cow pastures called Cowbot (Fig. \ref{fig:cowbot_profile}). Cowbot has been developed as part of a project titled `Agricultural Weed Control using Autonomous Mowers' funded by MN LCCMR. The project is 
a collaboration between University of Minnesota researchers and the Toro Company. In this paper we give an overview of the system design and present the navigation stack on the Cowbot. Cowbot has been designed to operate in the rugged environments on cow pastures and provides a cost-effective method for weed control in organic fields.


The density of weeds on a pasture affects the cost of operation of the Cowbot. In fields with dense weed cover, a boustrophedon coverage path \cite{cpp} is usually the most efficient path to mow all weeds. In case of low density or localized distribution of weeds, a boustrophedon path results in considerable wastage of resources, specially on large fields. We assume a finite field of view of the Cowbot within which it is able to detect and localize all weeds. Then, an intuitive solution is to do a random search on the pasture and mow weeds as they are detected. This strategy would eventually allow the Cowbot to mow all weeds but as shown in Section \ref{ssec:sim_results}, it does not lead to major cost savings since the random search results in repetitive coverage of same areas. Another major drawback to any random search based method is not knowing when to terminate. Common termination criteria include stopping when no new weeds are detected for a period of time or when a threshold for maximum path length (or time) is reached. This necessitates the design of intelligent path planning algorithms to ensure all weeds on the pasture are detected and mowed by the autonomous mower. We refer to this problem as the mower routing problem (MRP). In this paper, we develop two online path planning algorithms, called JUMP and SNAKE, to efficiently solve MRP. Our path planning algorithms are able to utilize new information about weeds on the pasture to improve path length by up to 60 \% as compared to boustrophedon and random-search based coverage paths. We implement our algorithms on the Cowbot and perform field experiments to validate that our algorithms are suitable for real-time path planning. We also perform large-scale computational simulations and benchmark the performance of our solution methods with online and offline baseline methods.






%


\section{Related Work}






The design and development of mobile robots for weed removal applications has witnessed growth in recent years. Bakker et. al. \cite{BAKKER201063} have designed a weeding robot for intra-row weed removal. Bawden et. al. \cite{bawdenAGBOT} develop a versatile agricultural robot equipped with a heterogeneous set of weeding implements and develop weed detection methods. Grimstad and From \cite{thorvald} present the design of Thorvald agricultural robot and demonstrate the robustness and modularity of their design in field trials. While there exists a body of literature on weed detection, and system design, there is relatively little work on path planning for agricultural robots beyond coverage path planning. Weed distribution on pastures is not known a priori and hard to estimate. As a result boustrophedon like coverage paths can lead to large wastage in resources. 

MRP finds relevance in many well-studied problems such as variants of the traveling salesman problem (TSP), area coverage and curvature constrained path planning. In dynamic TSP \cite{dtsp_review} for instance, the distribution of city locations is unknown at the start of the problem. However, unlike in MRP wherein the mower must search for weeds on the pasture, city locations in DTSP are generated by a stochastic process and made available to the salesman during path execution. Another variant of TSP that requires the path to satisfy non-holonomic constraints of a Dubins motion model \cite{dubins} is called Dubins TSP \cite{rathinam2007nonholonomic}. While both of these problems find similarity to MRP, they do not capture all of the constraints and hence their solution methods cannot be used to solve MRP. Coverage path planning (CPP) \cite{cpp} also relates to MRP and addresses coverage of known spaces with a finite sensor footprint. Choset \cite{cpp} presents the boustrophedon cellular decomposition method to find solutions to CPP. Variants of the problem include energy \cite{Yu2019CoverageOA} and curvature constrained \cite{ag_coverage} \cite{dcov} coverage planning. However, CPP and its variants address the problem in a static environment and do not consider online planning to visit points of interest within the environment. Another closely related line of research is the design of random-search based path planning algorithms like rapidly-exploring random trees (RRT) and its variants. While RRT is an efficient solution for path planning with obstacle avoidance, random-search based methods are not best suited for coverage and TSP-like problems such as MRP.

There is no existing work in the literature that addresses the mower routing problem. In this paper, we formalize the problem statement for MRP and develop two online solution methods, called JUMP and SNAKE, to plan efficient paths for the mower. We compare the performance of our online planning methods with coverage and TSP based solution methods. We also compare our methods with a random-search based reactive online planner. We show that our algorithms lead to up to 60\% shorter paths as compared to boustrophedon or random-search based methods by efficiently utilizing information about weed locations. We deploy our algorithms on the Cowbot and conduct proof-of-concept experiments to verify computational efficiency and suitability for real-time path planning.  

\subsubsection*{Paper Organisation} We present the system design for Cowbot in Section \ref{sec:robot_design} and formalize the problem statement for MRP in Section \ref{sec:problem_formulation}. In Section \ref{sec:preliminaries}, we discuss preliminaries, and upper and lower bounds to the MRP. We present our solution methods for MRP in Section \ref{sec:sol_methods}. In sections \ref{ssec:sim_results} and \ref{sec:exp_results}, we present results from simulations and field experiments, respectively. In section \ref{sec:concFut} we discuss conclusions and identify future directions to this work.

\section{Cowbot System Design}\label{sec:robot_design}
Cowbot is built on the Toro Groundmaster 3280-D rotary mower platform. It has a four-wheel drive transmission system with steering control on the rear wheels and a turning radius of 1.5 m at the geometric center. It weighs around 1800 lbs. It is customized to operate with a 32 kW electric motor powered by a 28.8 kWh lithium ion battery pack. It is designed to be a standalone solution for weed control. As part of the project, a cargo van was equipped with retractable solar panels on the roof and climbing ramps to house the Cowbot and recharge it on the pasture without the need for any charging infrastructure. Cowbot incorporates a suite of safety features including a wireless remote emergency stop button, pressure activated perimeter bumpers and emergency stop knobs on the front and back of the mower. It comprises a flail-deck weeding implement to reduce the risk of flying debris during autonomous operations. The flail-deck has a cutting width of 1.3 m and weighs 500 lbs. It is custom mounted with high ground clearance to achieve a cut-height of 8 in. The high cut ensures sufficient volume of grass for the cows even after mowing while also stunting weed growth.

 
Cowbot uses a GPS-based point to point inertial navigation system and is equipped with two multi-band RTK GNSS receiver units (SwiftNav Duro Inertial) with inbuilt IMU sensors to provide high accuracy location and heading estimates. It is also equipped with front facing imaging sensors that includes two RGB-D cameras (Intel RealSense D455) and a lidar sensor (Velodyne Puck VLP-16). The perception stack of the Cowbot that aims to build capability for visual inertial navigation and weed detection on the field is currently under development. The sensors connect to an onboard laptop computer (Dell Precision 7530) that runs the navigation software. The navigation software on Cowbot has a three-layer architecture. The top and middle layers execute on the laptop and the bottom layer executes on a programmable logic controller (TORO PLC model \# - Tec 5004) that controls the traction and steering systems. The top layer is the perception and planning layer that takes input from the user and on-board sensors, and plans paths for the mower. The middle layer comprises of high level control modules for point to point navigation. It communicates with the top layer over the robot operating system (ROS) and takes as input the waypoint path computed by the planner. It generates steering and speed commands for the bottom layer. The bottom layer generates signals for the electric motor and the hydraulic steering to follow commanded speed and steering values. The middle and bottom layers communicate using CANBUS messaging protocol.




\section{Problem Formulation}\label{sec:problem_formulation}

In this section, we define the system model for the Cowbot used in path planning and present the problem statement.

\subsection{System Model} \label{sec:system_model}

\begin{figure}
    \centering
    \subfloat[Robot Model]{\includegraphics[height = 2.5cm]{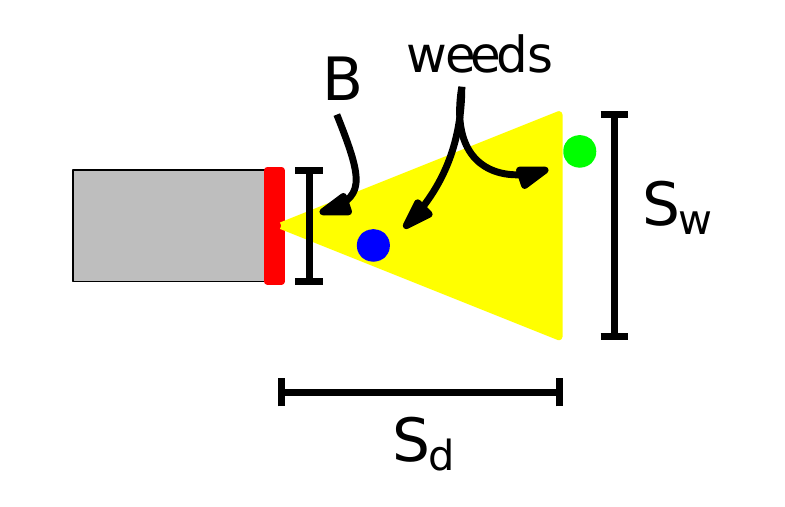}\label{fig:robot_model}}
    \subfloat[Cowbot on a hill]{\includegraphics[height = 2.5cm]{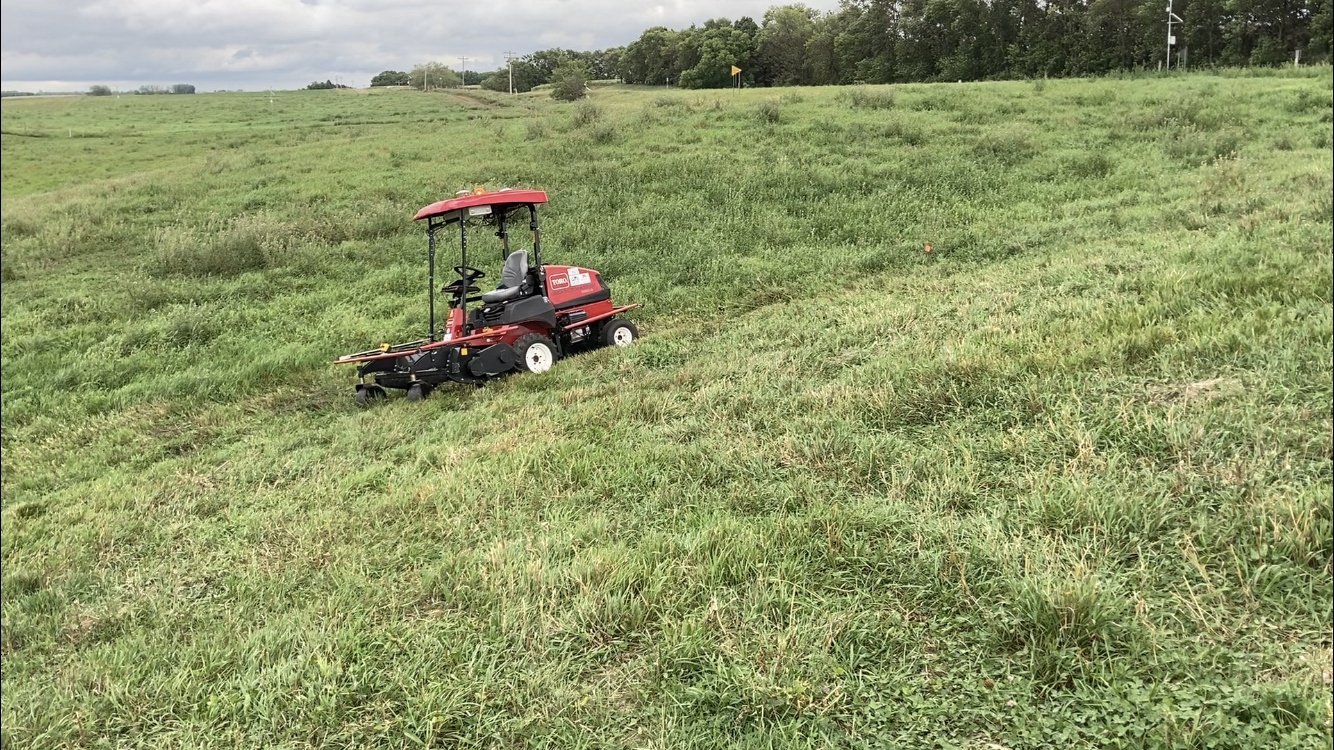}\label{fig:slope_cowbot}}
    \caption{ (a). Robot model showing the robot (gray), detection field of view (yellow), weeding implement (red) and weeds (solid disks) in the pasture. The green disk shows an undetected weed on the pasture and blue disk shows a weed detected by the mower within the detection field of view. (b). Cowbot operating on steep slopes in the pasture.}\label{fig:robot_model_pasture}
\end{figure}

As the Cowbot moves in the pasture, it sweeps the ground with the weeding implement and mows any weeds that it passes over. For the purpose of path planning, we approximate its motion model with non-holonomic motion constraints that follow Dubins car-like vehicle model \cite{dubins}. It has a finite turning radius, $\mathcal{R}$, and similar to the Cowbot is equipped with a front mounted weeding implement (see Figure \ref{fig:robot_model_pasture}) that it pushes as it moves around in the pasture. The weeding implement has a finite width, $\mathcal{B}$. A weed is considered mowed when the implement passes over it. 



We also consider the mower model to consist of a weed detection module that includes a suite of front facing imaging sensors. The weed detection field of view (FOV) is finite, triangular shaped and apexed at the front center of the mower (see Figure \ref{fig:robot_model} for an illustration of the FOV). The sensing depth range, $\mathcal{S}_{d}$, and width at the farthest point in the range, $\mathcal{S}_{w}$, define the range of the detection FOV. The mower is able to detect and localize a weed in the global frame of reference instantaneously as the weed enters its FOV.


\subsection{Problem Statement}
Let $\mathcal{E}$ be a rectangular pasture. The mower does not have any prior information on the location or distribution of weeds within $\mathcal{E}$. To ensure all weeds on the pasture are detected, the mower must cover the entire pasture with its FOV and perform path planning to visit and mow each of the detected weeds. The \emph{Mower Routing Problem (MRP)} can then be defined as follows:

Given a pasture $\mathcal{E}$ of unknown weed distribution and an autonomous mower equipped with a front mounted weeding implement and a finite FOV (as defined in Section \ref{sec:system_model}), develop an online path planning policy that plans a path for the mower to detect and mow all weeds in $\mathcal{E}$ while minimizing the total length of the path.

\section{Preliminaries}\label{sec:preliminaries}

In this section, we define notation used to develop the solution methods, discuss Dubins path preliminaries and compute bounds on the solution space to MRP.


Given a rectangular pasture $\mathcal{E}$ of size ($L\times W$), we consider an inertial frame of reference such that the four corners of the pasture are located at $(0,0), (L,0), (L,W)$ and $(0,W)$. For ease of exposition we will refer to locations with bigger $y$-coordinates as being located above locations with smaller $y$-coordinates, and vice-versa as being located below. Current state of the mower is represented as $(x_m,y_m,\theta_m)$, where $(x_m,y_m)$ are the location coordinates and $\theta_m$ is the heading angle. The mower maintains a list, $\mathcal{W}$, of weeds detected on the pasture that are yet to be mowed. 

\subsection{Dubins Paths}\label{sec:dubins}
We use primitives from Dubins shortest path algorithm for curvature constrained vehicles \cite{dubins} to compute traversable paths for the Cowbot. Specifically we consider Dubins paths of the type $C^+SC^-$, where $C^+$ and $C^-$ are circular arcs in opposite direction and $S$ is a straight line path. LSR and RSL paths are realizations of the $C^+SC^-$ path type, where R refers to a circular arc in the clockwise direction and L refers to a circular arc in the counter clockwise direction. Both circular arcs turn with the minimum turn radius, $\mathcal{R}$.

\subsection{Bounding the Solution Space} \label{sec:sol_bounds}

Weed distribution on the pasture significantly affects the length of the mower's path needed to mow all weeds.  We present two edge-case scenarios and use them to compute upper and lower bounds on the solution space.

First, consider a pasture with a uniform weed distribution and densely packed weeds. The mower must mow every point on the pasture to ensure that all weeds are mowed. An optimal mower path in this scenario is a boustrophedon coverage path (BCP). A boustrophedon path \cite{cpp} to cover the pasture comprises of straight line passes of length $L$ across the pasture that are parallel to $x$-axis. Consecutive passes in the boustrophedon path have opposite direction of motion and are $\mathcal{B}$ distance away from each other. Let $y_p(i)$ be the $y$-coordinate corresponding to the $i^{th}$ pass in the path and $\theta_p(i)$ be the direction of motion on the pass. The first pass of the path is along the line, $y_p=\mathcal{B}/2$ and the last pass is along the line $y_p=\lceil{\frac{W}{\mathcal{B}}}\rceil \times \mathcal{B}  - \mathcal{B}/2$. When it is clear from the context we use $y_p$ and $\theta_p$ interchangeably with $y_p(i)$ and $\theta_p(i)$, respectively. Such a path ensures all weeds on the pasture are mowed regardless of the weed distribution and provides an upper bound on the path length for the mower.
Second, consider the scenario with no weeds on the pasture. Since the mower does not have any information about the weed distribution, it must at least cover the pasture with its FOV to search for weeds. Consider another BCP but with spacing between consecutive passes equal to $\mathcal{S}_w$. This reduces the total number of passes by a factor of $\frac{\mathcal{S}_w}{\mathcal{B}}$. At the end of this path, the mower has deterministic knowledge of all weed locations (or of the absence of weeds) on the pasture. This is the minimum length path needed for the mower to search for weeds on the pasture and in the case no weeds are detected it forms the solution to the MRP instance. This provides a lower bound on the solution space for MRP. 
\section{Solution Methods}\label{sec:sol_methods}

In this section, we present solution methods to the mower routing problem to detect and mow weeds on a pasture. We design two online planning algorithms, named JUMP and SNAKE, that build on top of the boustrophedon coverage path. We present systematic modifications to the boustrophedon path that account for Cowbot's curvature constraints and utilize the width of the weeding implement to enable efficient online coverage of the pasture.





\begin{algorithm}
\caption{JUMP-planner ($L,W,\mathcal{S}_d,\mathcal{S}_w,\mathcal{B},\mathcal{R}$)}\label{algo:jumpAlgo}
\begin{algorithmic}[1]
\State $y_p = \mathcal{B}$/2, $\theta_p = 0$,  $\mathcal{W}$ = [ ]
\State mission\_completed = False 
\State jump\_flag = 0 
\State current\_pose =  \{$0,~y_p,~\theta_p$\}
\While{mission\_completed is False}
\If{$\theta_p == 0$}\Comment{$\mathcal{I}$: vector of 1s}
\State current\_path = [0 : $L$, $y_p.\mathcal{I}$, $\theta_p.\mathcal{I}$] 
\ElsIf{$\theta_p == \pi$}
\State current\_path = [$L$ : 0, $y_p.\mathcal{I}$, $\theta_p.\mathcal{I}$]
\EndIf
\State transit\_path = dubins(current\_pose, current\_path[0])
\State current\_path.prepend(transit\_path)
\While{not reached end of current\_path}
\State add new weeds detected in current FOV to $\mathcal{W}$
\If{jump\_flag == 1}
\State take next step on current\_path
\State remove mowed weeds from $\mathcal{W}$
\If{jump is complete}
\State jump\_flag = 0
\EndIf
\State continue
\EndIf
\State $\Delta$ = find\_available\_jump()
\If{$\Delta$ is not empty and $\Delta$ is feasible}
\If{$\theta_p$ == 0}
\State delete steps in current\_path with $x \leq x_e$
\ElsIf{$\theta_p == \pi$}
\State delete steps in current\_path with $x \geq x_e$
\EndIf
\State current\_path.prepend($\Delta$)
\State jump\_flag = 1
\EndIf
\State take next step on current\_path
\State remove mowed weeds from $\mathcal{W}$
\EndWhile
\If{termination criteria is true}
\State mission\_completed = True
\State \Return
\EndIf
\State Compute $y_p$ for next pass
\State $\theta_p = \pi - \theta_p$
\EndWhile
\end{algorithmic}
\end{algorithm}

\subsection{JUMP Online Planning Algorithm}




The JUMP planner modifies a BCP path by introducing the ideas of \emph{jumps} and \emph{spring}. It computes a path for the mower that starts at the bottom of the pasture and makes its way towards the top while doing to-and-fro passes across the pasture much like a boustrophedon path. A \emph{jump} allows the mower to make detours to mow nearby weeds within its FOV and then return back to the current pass to continue on its path. \emph{Spring} behavior of the JUMP planner permits variable spacing between consecutive passes, unlike a boustrophedon path, resulting in a stretch and shrink pattern to better utilize new information about weeds on the pasture. We present the pseudocode for the JUMP planner in Algorithm \ref{algo:jumpAlgo}.



\begin{figure}[h]
\centering
\includegraphics[scale = .8]{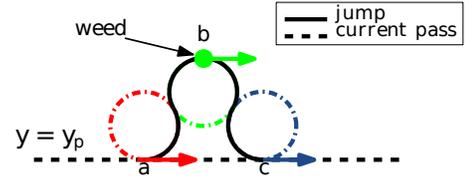}
\caption{LSR-RSL jump to mow a weed at point \textbf{b} with $\theta_p = 0$. \textbf{a} and \textbf{c} are the start and end points of the jump.}\label{fig:detour_jump}
\end{figure}

When the mower determines that there are no weeds directly in front on its current path, it makes a search for nearby weeds that it can mow by making a quick detour. We restrict these detours to weeds that are located above the current pass, thus leading to the name \emph{jump}. A jump consists of two sub-paths: first, to go \emph{up} to the weed's location from the current position and second, to come back \emph{down} to the current pass. We use Dubins paths of type RSL and LSR to compute sub-paths for a jump. Each jump is defined by three waypoints: start and end points of the jump and the location of the weed. We fix the heading angle at each of the three waypoints to $\theta_p$. When $\theta_p = 0$, jumps comprise of sub-path LSR to go to the weed and RSL to come back to the path (Figure \ref{fig:detour_jump}), and vice-versa when $\theta_p = \pi$. Ties among jumps starting at the same point are broken arbitrarily.

At the end of a pass, the planner must compute the $y_p$ value for the next pass with $\theta_p = \pi - \theta_p$. JUMP planner admits dynamic spacing between consecutive passes allowing a \emph{spring}-like behavior in the paths planned. 
$y_p(i+1)$ is computed as follows:
\begin{align}
    y_p(i+1) = \rm{min}
    \begin{cases}
    \min\limits_{\forall i:w_i \in \mathcal{W}} y_i+\mathcal{B}/2\\
    y_p(i)+\mathcal{S}_w/2\\
    W-\mathcal{B}/2
    \end{cases}\label{eq:jumplow_yp}
\end{align}
JUMP planner allows for ($y_p(i+1) < y_p(i)$) to utilize any new information about weeds on the pasture. The mower continues traversing on consecutive passes across the pasture, alternating its direction of motion, until the termination criteria for JUMP algorithm is met. The termination criteria, evaluated at the end of each pass, is as follows:
\begin{align}
(y_p == W-\mathcal{B}/2) \cap (\mathcal{W} == \phi)\label{eq:jump_terminate}
\end{align}

\subsection{SNAKE Online Planning Algorithm}
The SNAKE planner derives its name from serpentine to-and-fro patterns across the pasture generated in paths computed by the planner. It builds on top of the JUMP planner by allowing detours in both directions and not requiring the mower to return to its original path after mowing a weed. Thus only completing the first sub-path on a jump. The SNAKE planner enforces a consistent direction of motion on a pass by fixing the heading angle at each weed location to be equal to $\theta_p$. When the mower completes a sub-path, it resumes the search for the next closest weed on the pasture along its direction of motion. If there does not exist a valid sub-path to a weed, the mower continues to move straight ahead with heading angle $\theta_m = \theta_p$.





For sub-path computation, we use primitives from the Dubins shortest path algorithm as discussed in Section \ref{sec:dubins}. If $\theta_p=0$ the SNAKE planner considers Dubins paths of type LSR and RSL to compute sub-paths to weeds located above and below $y_m$ respectively, and vice-versa when $\theta_p=\pi$. While the mower is executing a sub-path, the planner does not search for new sub-paths. When the mower reaches the end of the pasture along its direction of motion, the planner computes $y_p$ value for the next pass with $\theta_p(i+1) = \pi - \theta_p(i)$. Similar to the JUMP algorithm, the SNAKE planner starts from the bottom of the pasture and moves towards the top in consecutive passes across the pasture. It computes the $y_p(i+1)$ value as follows:
\begin{align}
    y_p(i+1) = \rm{min}
    \begin{cases}
    y_p(i)+\mathcal{S}_w/2 + \mathcal{B}/2\\
    W - \mathcal{B}/2
    \end{cases}\label{eq:yp_compute_snake_fixed}
\end{align}
Termination criteria for the SNAKE planner, evaluated at the end of each pass, is:
\begin{align}
    (y_m \geq W - \mathcal{B})~ \cap ~(y_p == W - \mathcal{B}/2)~ \cap ~(\mathcal{W} == \phi)\label{eq:terminate_snake_fixed}
\end{align}

We present pseudocode for the SNAKE planner in Algorithm \ref{algo:snakeAlgo}. We also define a variation of the SNAKE planner called RESTRICTED-SNAKE (R-SNAKE) that trades-off weed coverage in favor of shorter path length for the mower. While the SNAKE planner admits sub-paths to all unmowed weeds along its direction of motion, R-SNAKE restricts its search to weeds, $w_i \equiv (x_i,y_i)$, located at most a fixed threshold below the $y_p$ value of the current pass i.e.:
\begin{align}
&y_i \geq y_p-\frac{3}{2}\mathcal{S}_w& \label{eq:snake_wf_lb}
\end{align}
R-SNAKE uses the following termination criteria evaluated at the end of each pass:
\begin{align}
    (y_m \geq W - \mathcal{B})~ \cap ~(y_p == W - \mathcal{B}/2)\label{eq:terminate_snake_static_limited}
\end{align}



\begin{algorithm}
\caption{SNAKE-planner ($L,W,\mathcal{S}_d,\mathcal{S}_w,\mathcal{B},\mathcal{R}$)}\label{algo:snakeAlgo}
\begin{algorithmic}[1]
\State $y_p = \mathcal{B}$/2, $\theta_p = 0$, $\mathcal{W}$ = [ ]
\State mission\_completed = False 
\State wriggle\_flag = 0 
\State current\_pose =  \{$0,~\mathcal{B}$/2, 0\}
\While{mission\_completed is False}
\If{$\theta_p == 0$}\Comment{$\mathcal{I}$: vector of 1s}
\State current\_path = [$0:L$, $y_p.\mathcal{I}$, $\theta_p.\mathcal{I}$]
\ElsIf{$\theta_p == \pi$}
\State current\_path = [$L:0$, $y_p.\mathcal{I}$, $\theta_p.\mathcal{I}$]
\EndIf
\State transit\_path = dubins(current\_pose, current\_path[0])
\State current\_path.prepend(transit\_path)
\While{$x_m < L$}
\State add new weeds detected in current FOV to $\mathcal{W}$
\If{wriggle\_flag == 1}
\State take next step on  current\_path
\State remove mowed weeds from $\mathcal{W}$
\If{reached end of current\_path}
\State wriggle\_flag = 0
\If{$\theta_p$ == 0}
\State current\_path = [$x_m:L$, $y_m.\mathcal{I}$, $\theta_p.\mathcal{I}$]
\ElsIf{$\theta_p == \pi$}
\State current\_path = [$x_m:0$, $y_m.\mathcal{I}$, $\theta_p.\mathcal{I}$]
\EndIf
\EndIf
\State continue
\EndIf
\State $\Omega$ = find\_valid\_subpath()
\If{$\Omega$ is not empty}
\State current\_path = $\Omega$
\State wriggle\_flag = 1
\EndIf
\State take next step on current\_path
\State remove mowed weeds from $\mathcal{W}$
\EndWhile
\If{termination condition is true}
\State mission\_completed = True
\State \Return
\EndIf
\State Compute $y_p$ for next pass
\State $\theta_p = \pi - \theta_p$
\EndWhile
\end{algorithmic}
\end{algorithm}

\section{Simulation Results}\label{ssec:sim_results}

To evaluate the performance of the online planning algorithms presented in Section \ref{sec:sol_methods} we designed a simulated pasture environment and an autonomous mower operating on the pasture. The simulator was implemented in Python 3 programming language and uses PyDubins and Elkai libraries to compute Dubins paths and TSP tours, respectively. The simulations were run on a desktop computer with AMD Ryzen 7 processor and 32GB RAM. 

\begin{figure}
\subfloat[]{\includegraphics[width = 0.24\linewidth]{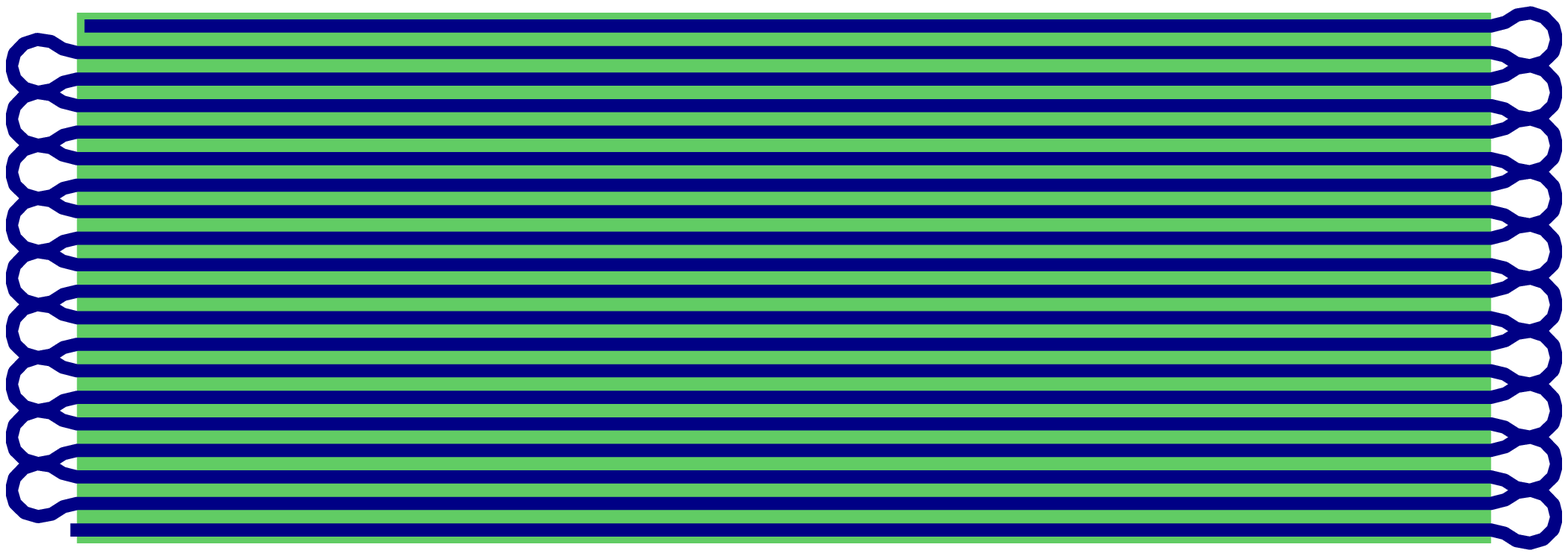}\label{fig:BCP}}\hspace{0.5mm}
\subfloat[]{\includegraphics[width = 0.23\linewidth]{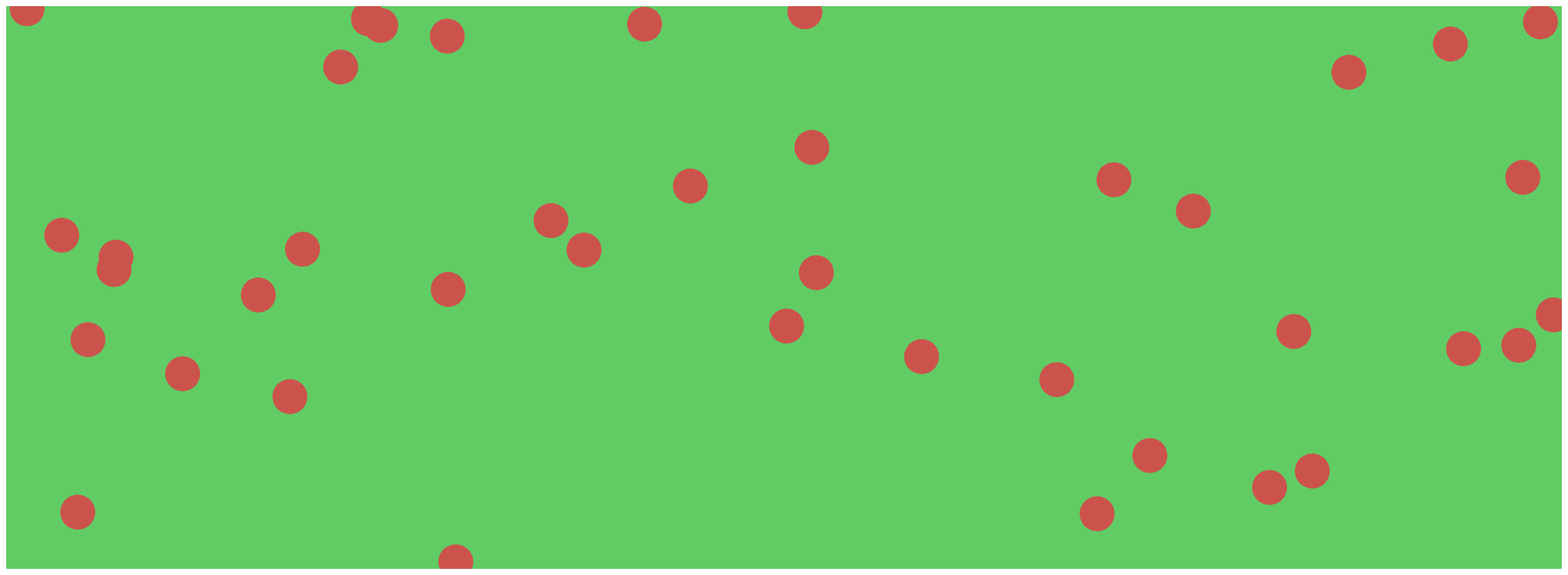}\label{fig:u40}}\hspace{0.5mm}
\subfloat[]{\includegraphics[width = 0.23\linewidth]{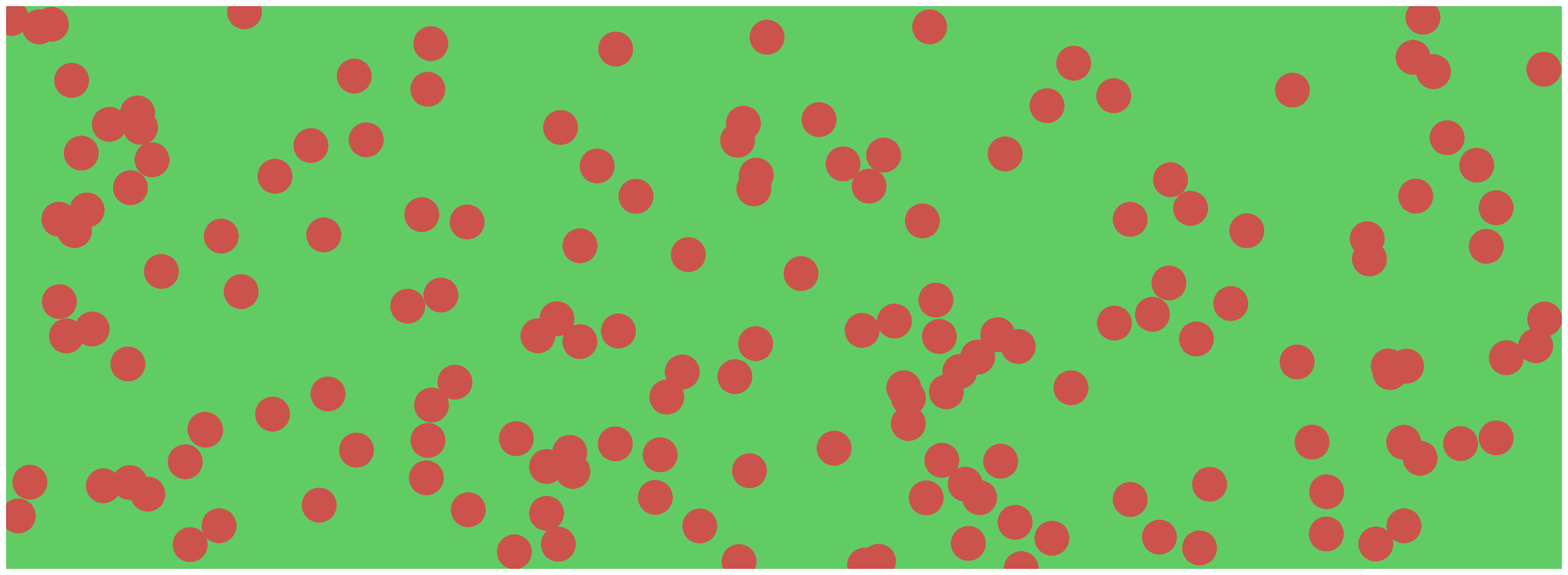}\label{fig:u160}}\hspace{0.5mm}
\subfloat[]{\includegraphics[width = 0.23\linewidth]{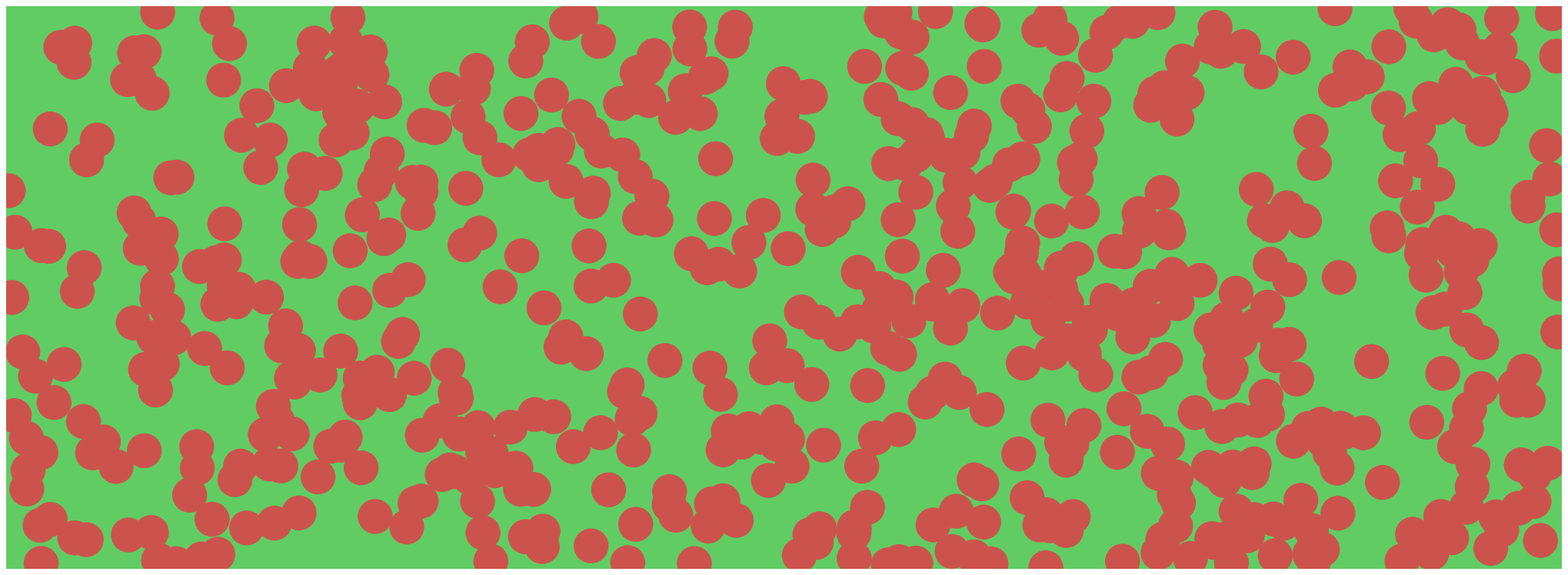}\label{fig:u640}}\\
\subfloat[]{\includegraphics[width = 0.24\linewidth]{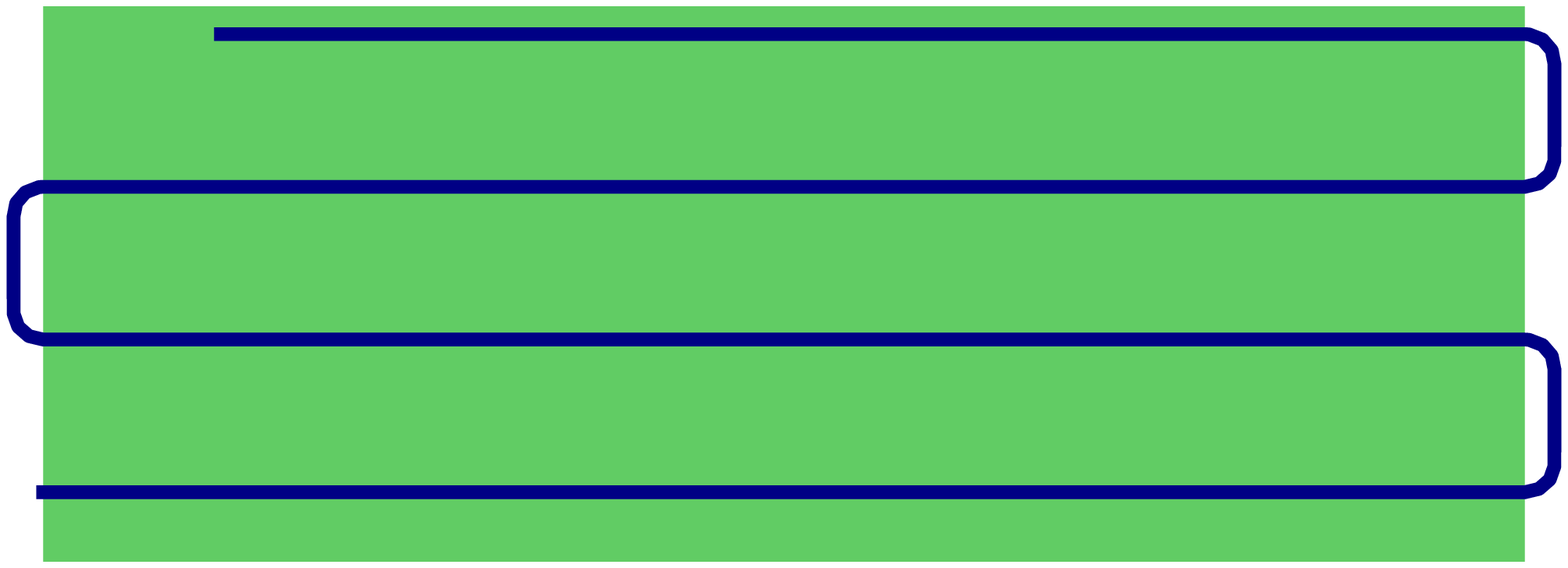}\label{fig:BCP_TSP}}\hspace{0.5mm}
\subfloat[]{\includegraphics[width = 0.23\linewidth]{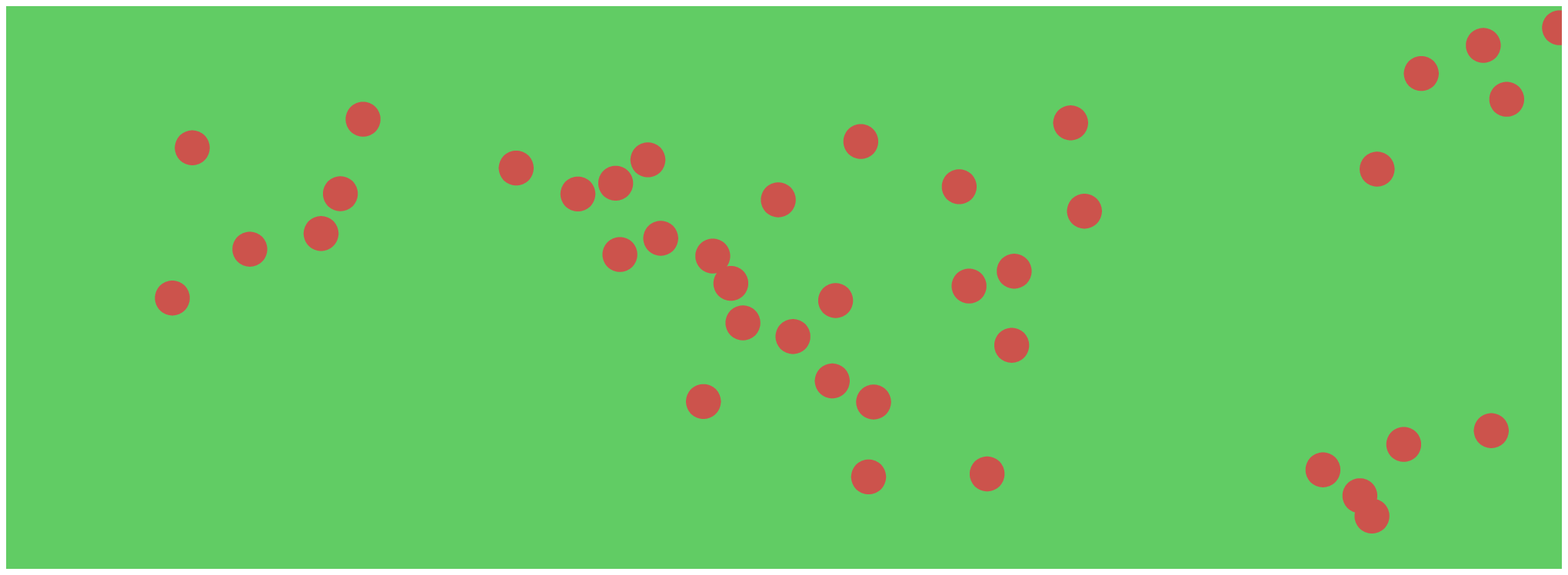}\label{fig:g40}}\hspace{0.5mm}
\subfloat[]{\includegraphics[width = 0.23\linewidth]{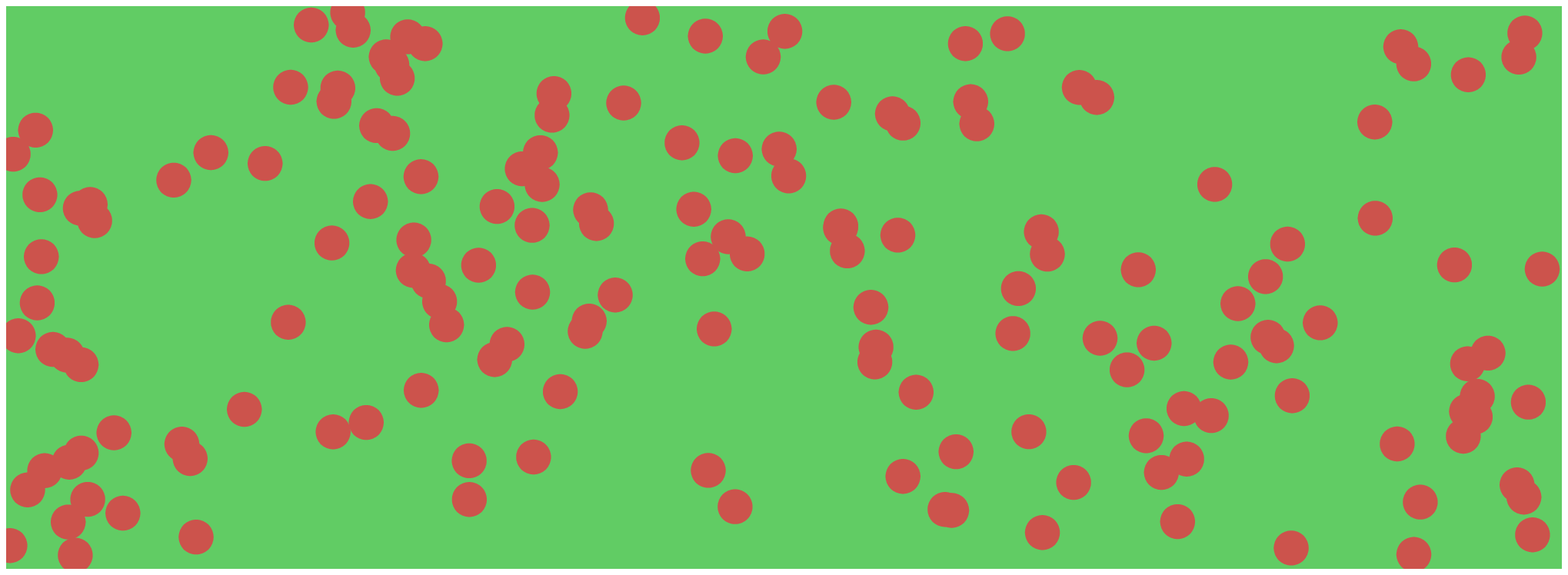}\label{fig:g160}}\hspace{0.5mm}
\subfloat[]{\includegraphics[width = 0.23\linewidth]{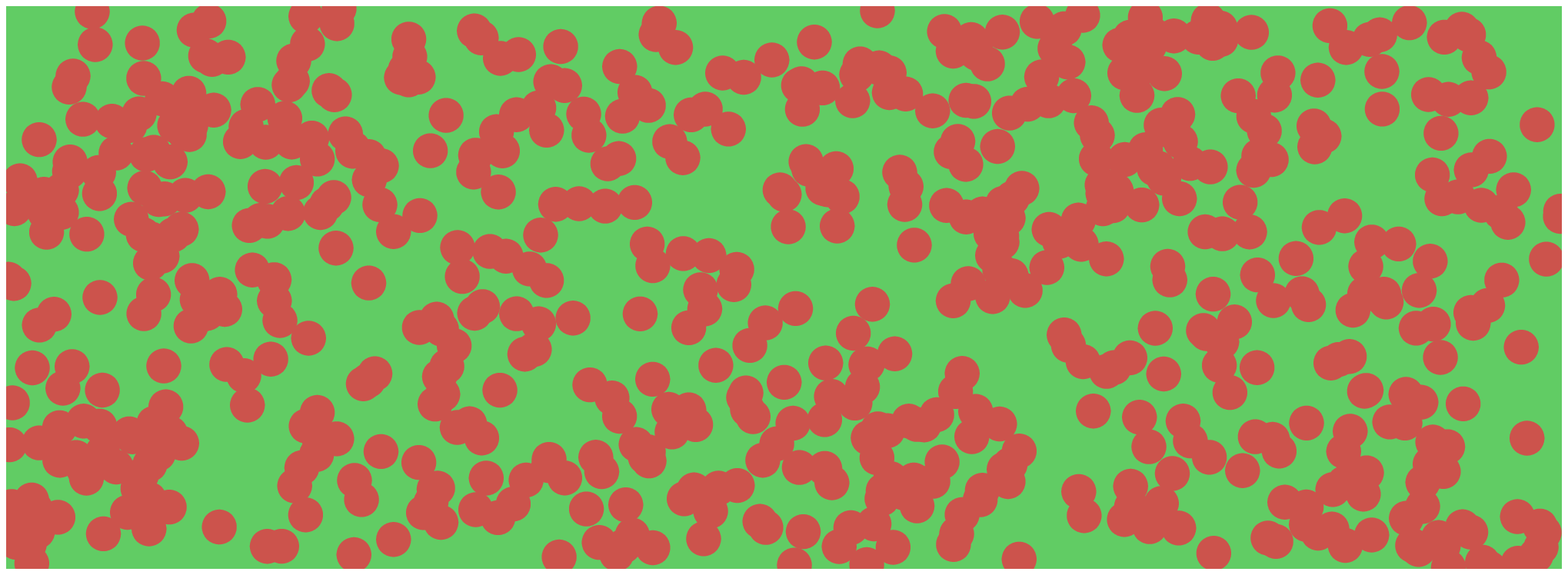}\label{fig:g640}}
\caption{(a). upper bounding BCP path. (e). lower bounding BCP path. (b)-(d) show uniform distribution of weeds on a simulated pasture with 40, 160 and 640 weeds respectively. (f)-(h) show Gaussian distribution of weeds on a simulated pasture with 40, 160 and 640 weeds respectively.}
\label{fig:simpastureplots}
\end{figure}
We compare the performance of our online algorithms with three baseline solutions. First, the boustrophedon coverage path (BCP) used to compute upper bounds on the path length for MRP as given in Section \ref{sec:sol_bounds}. We recall that path length in case of BCP is independent of the weed density and spatial distribution and is determined by the size of the pasture and mower's parameters. The second baseline called BCP-TSP comprises of two phases. The first phase performs visual coverage of the pasture using the lower bounding BCP as described in Section \ref{sec:sol_bounds}. The second phase computes a TSP tour for the Cowbot to mow weeds detected in the first phase. The third baseline is an online reactive planner (REACT) that performs a random search for weeds on the pasture and mows weeds in the order they are detected. The planner generates random waypoints for the mower to perform a search. When it detects a weed, the mower aborts its search and starts to visit weed locations in FIFO order. It resumes the search once all weeds are mowed. It terminates when the path length equals BCP path length (upper bound).

\subsection{Simulation Setup} \label{sec:sim_setup}
We perform the simulations in a pasture of size 100m $\times$ 40m. To populate weeds on the pasture we use two different spatial distributions, uniform ($U$) and Gaussian ($G$). In the case of Gaussian distribution, 20\% of the weeds are sampled uniformly at random and chosen as seeds for Gaussian clusters used to generate rest of the weed locations. Figure \ref{fig:simpastureplots} shows simulation instances with uniform (\ref{fig:u40}, \ref{fig:u160} and \ref{fig:u640}) and Gaussian (\ref{fig:g40}, \ref{fig:g160} and \ref{fig:g640}) weed distribution and paths planned by the different planners. The mower is configured to operate with a constant speed of 1 m/s and comprises of a front mounted weeding implement of width $\mathcal{B} = 2m$. We vary the number of weeds $\mathcal{N} = ||\mathcal{W}||$ on the pasture (in turn the weed density), the detection FOV and turning radius of the mower to study their effect on the length of the path computed by the online planners. For each combination of weed distribution ($U/G$) and weed count $\mathcal{N}$, we generate 100 random instances. Table \ref{tab:simParams} lists the names and range of values for each of the simulation parameters.

\begin{table}[]
\begin{center}
\resizebox{\linewidth}{!}{%
\begin{tabular}{|c|c|}
\hline
\textbf{Parameter} & \textbf{Values} \\ \hline
minimum turn radius: $\mathcal{R}$(m) &\{1,2,3,4,5\}  \\ \hline
depth of detection FOV: $\mathcal{S}_d(m)$ &\{4,8,12,16,20\} \\ \hline
width of detection FOV: $\mathcal{S}_w(m)$ &\{4,8,12,16,20\}  \\ \hline
\# of weeds: $\mathcal{N}$ &\{20,40,80,160,320,640\}\\ \hline
\end{tabular}}
\caption{Range of values of various simulation parameters.}
\label{tab:simParams}
\end{center}
\end{table}


\begin{figure}
\centering
\subfloat[JUMP]{\includegraphics[width = 0.33\linewidth]{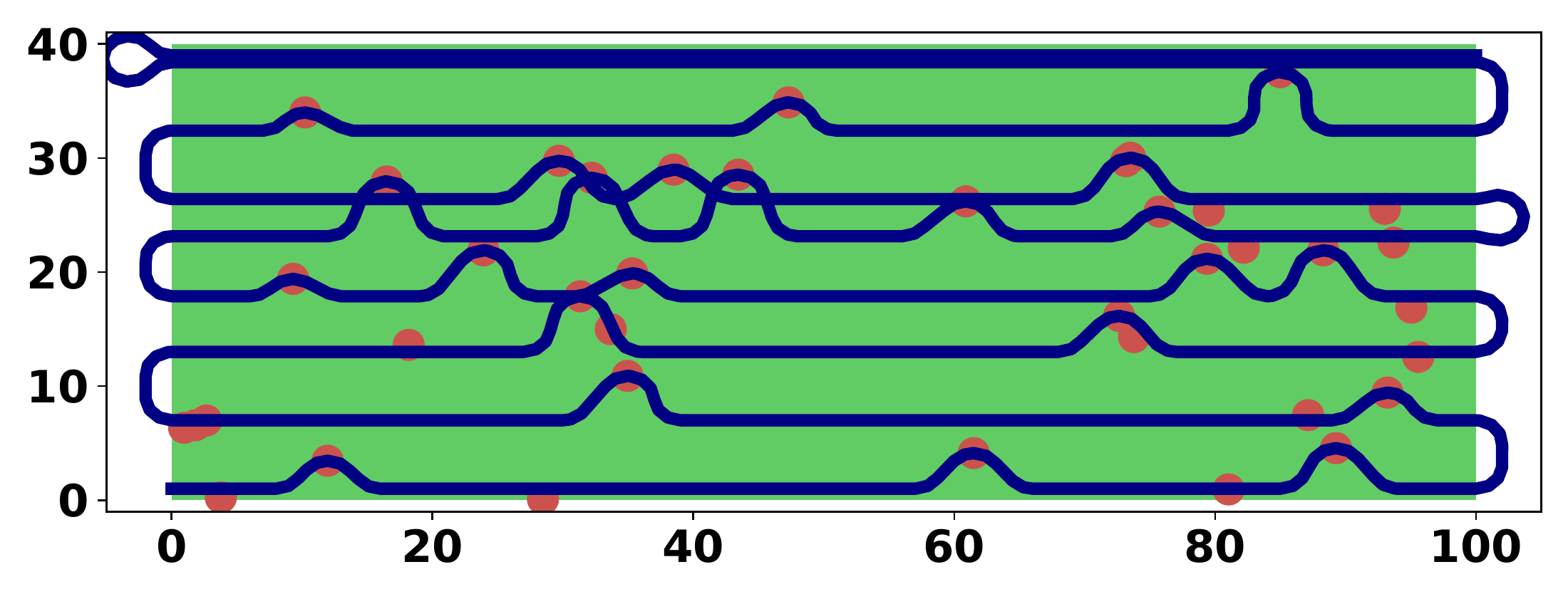}\label{fig:jumppath}}
\subfloat[SNAKE]{\includegraphics[width = 0.33\linewidth]{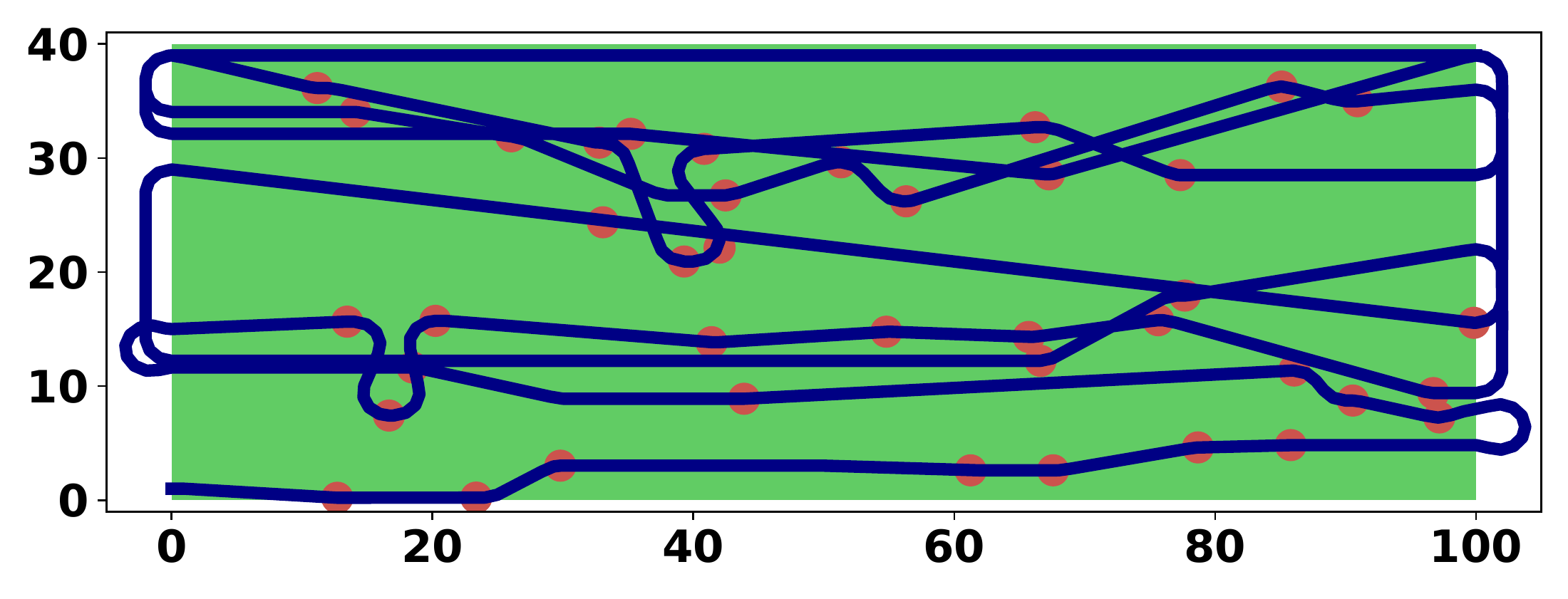}\label{fig:snakepath}}
\subfloat[R-SNAKE]{\includegraphics[width = 0.33\linewidth]{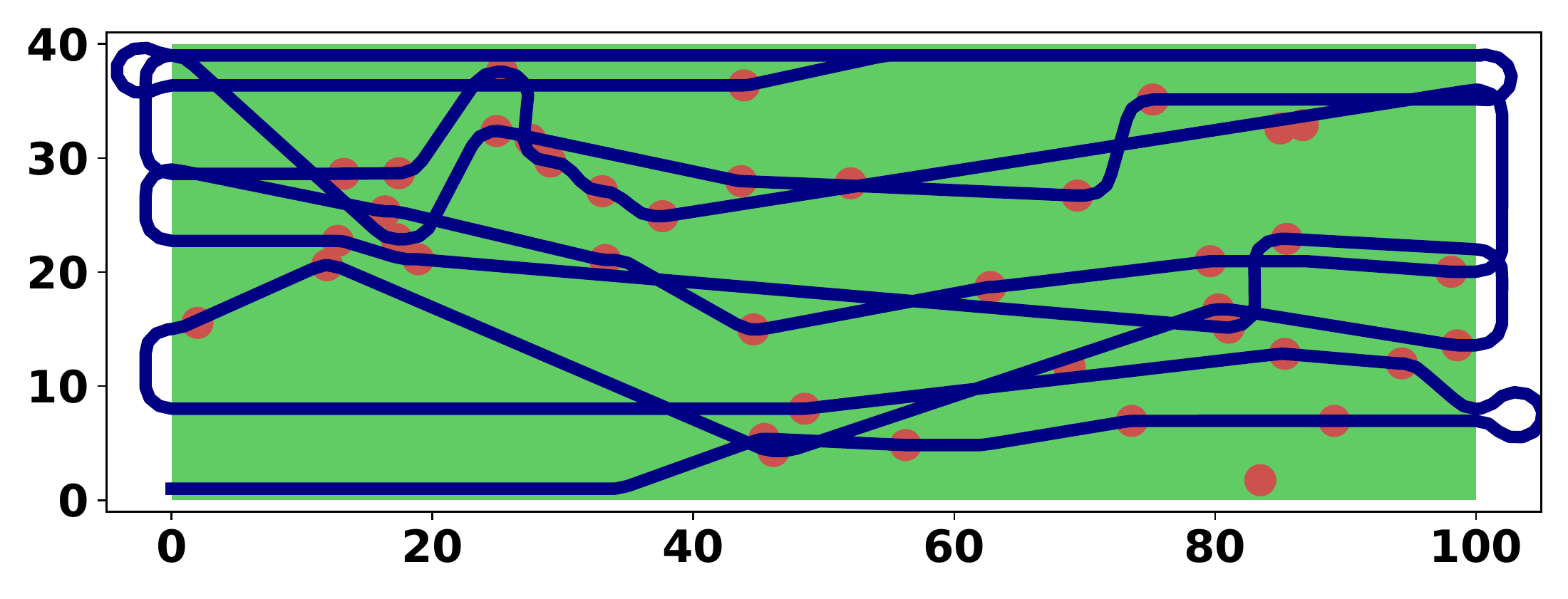}\label{fig:rsnakepath}}\\\vspace{-1mm}
\subfloat[BCP]{\includegraphics[width = 0.33\linewidth]{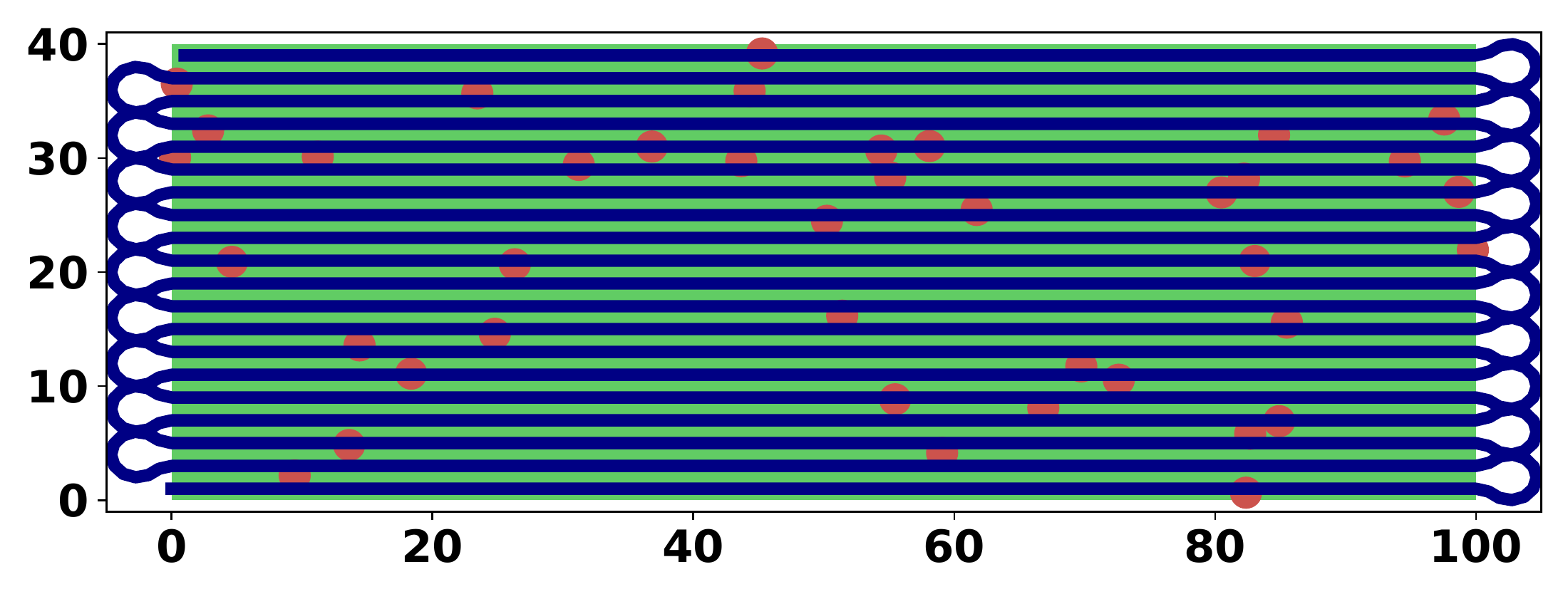}\label{fig:bcppath}}
\subfloat[BCP-TSP]{\includegraphics[width = 0.33\linewidth]{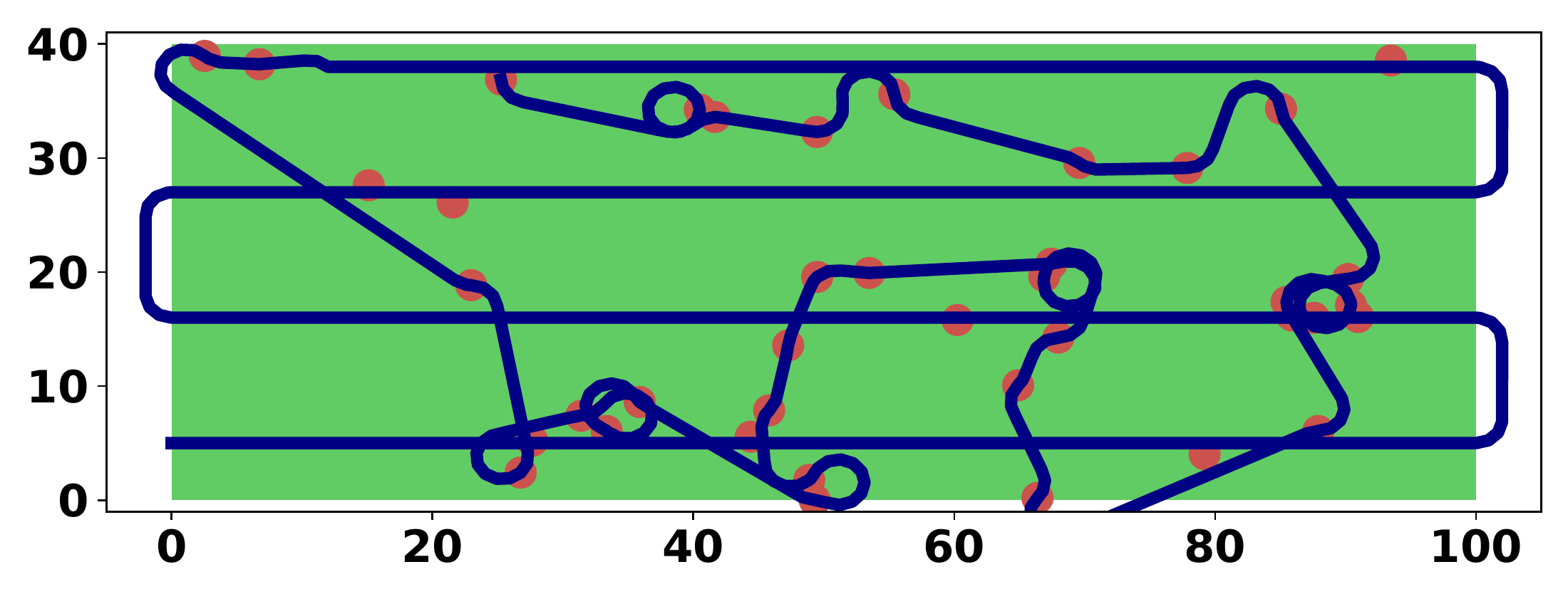}\label{fig:fovtsppath}}
\subfloat[REACT]{\includegraphics[width = 0.33\linewidth]{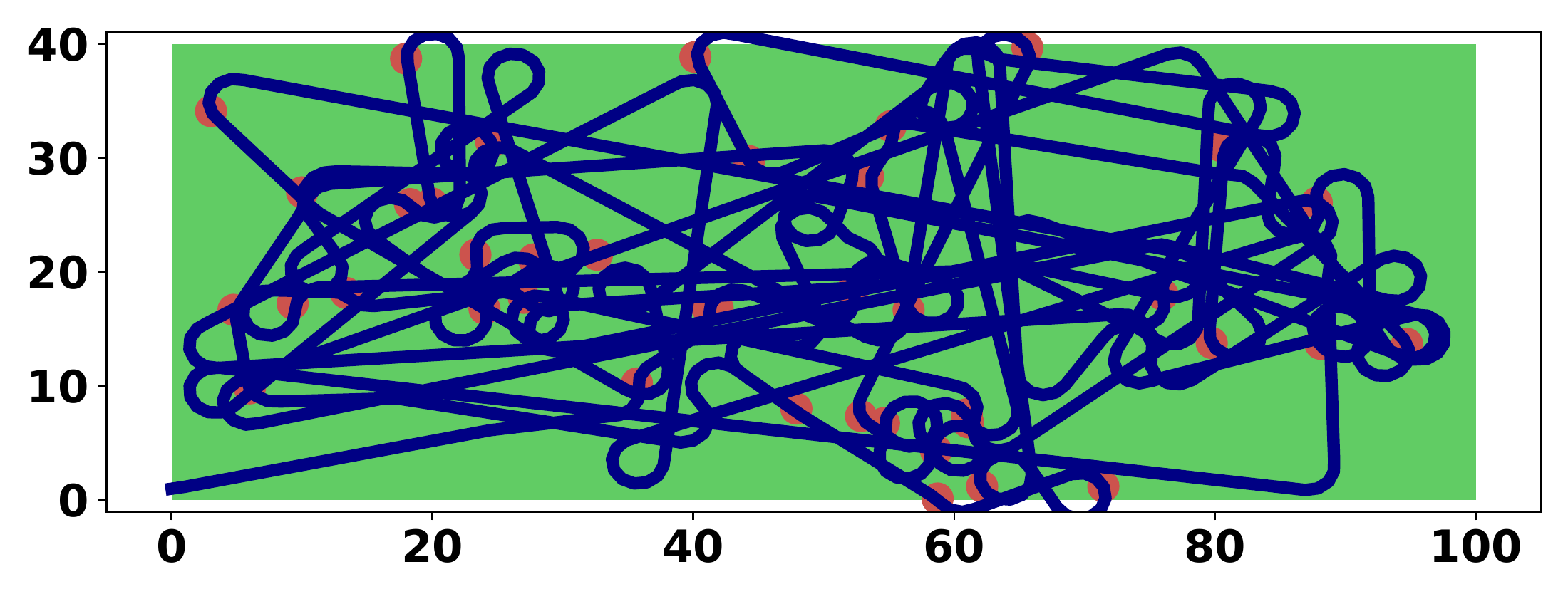}\label{fig:reactpath}}
\caption{Sample paths for the Cowbot planned using various planning algorithms to mow 40 weeds on a simulated pasture. Green area marks the pasture boundary, red disks mark the weeds and the solid blue line shows the mower path.}
\label{fig:simpathplots}
\end{figure}

\subsection{Results}


We use two evaluation metrics to compare and evaluate the performance of our solution methods - a). length of the mower's path and b). percentage of weeds mowed on the pasture. For ease of exposition we express the path length as a percentage of BCP path length that provides an upper bound on the solution space for MRP. 

\begin{figure}
\centering
\subfloat[]{\includegraphics[width = 0.5\linewidth]{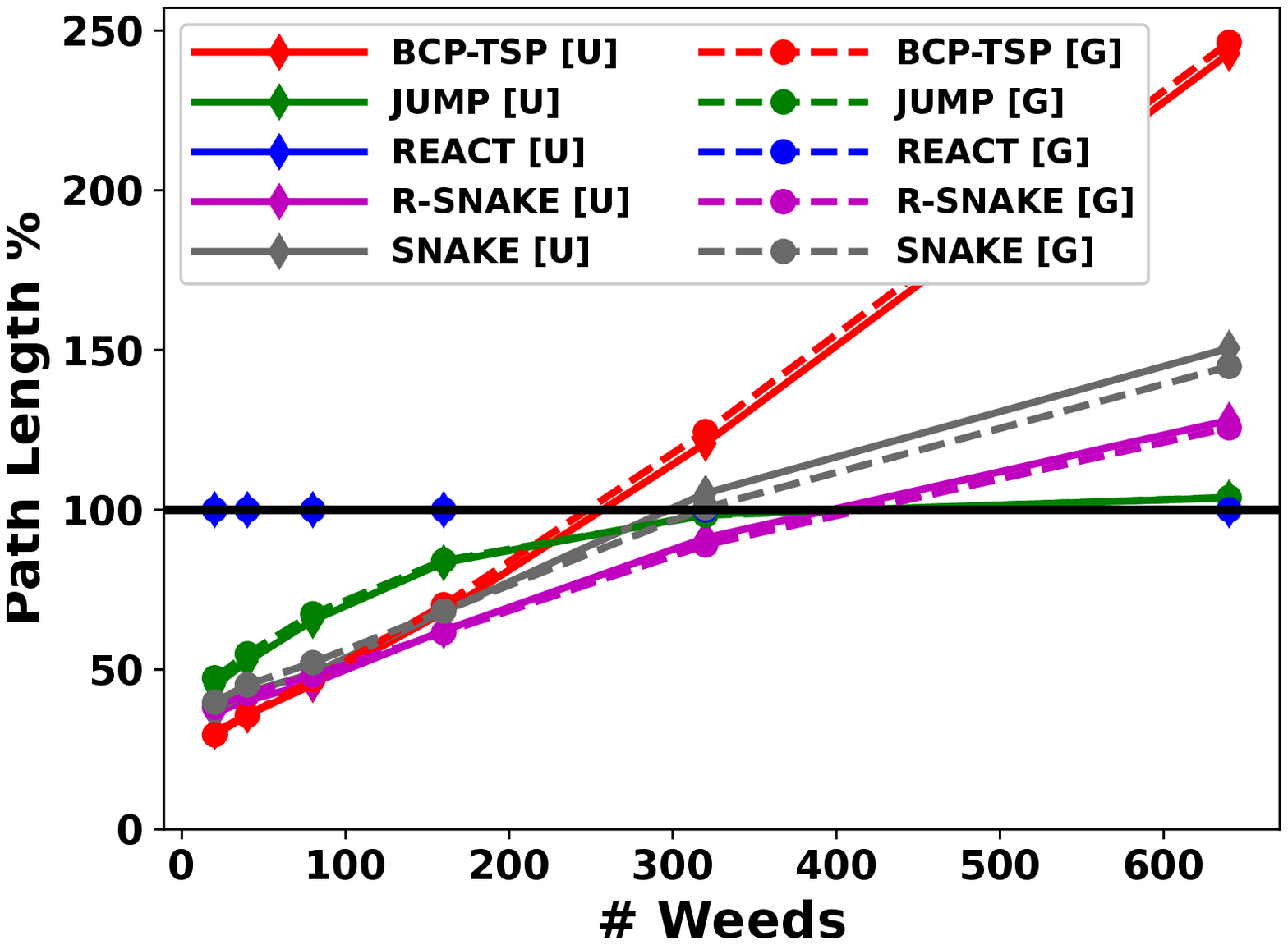}\label{fig:length_vs_weeds}}
\subfloat[]{\includegraphics[width = 0.5\linewidth]{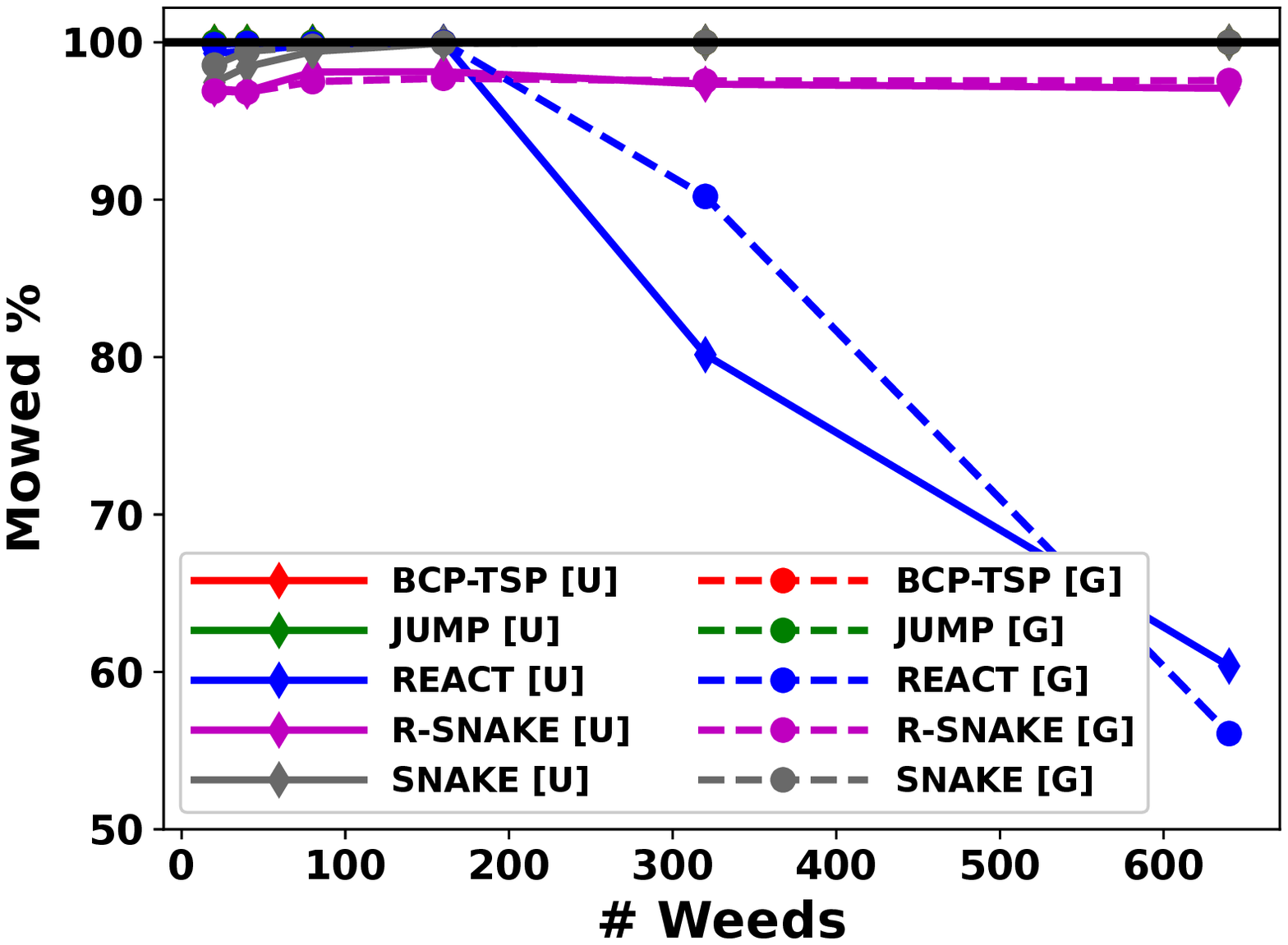}\label{fig:mowed_vs_weeds}} 
\caption{The plots show trends in (a) path length and (b) \% of weeds mowed, as a function of number of weeds on the pasture for various planning algorithms. Both plots show trends for both uniform and Gaussian distributions. Values for other parameters: $\mathcal{R}= 2m, \mathcal{S}_d = 12m, \mathcal{S}_w=12m$.}
\end{figure}

\subsubsection*{Effect of weed density} We observe that the density of weeds on the pasture visibly affects the path length (Fig. \ref{fig:length_vs_weeds}); it increases consistently with increase in the number of weeds for all planning algorithms. For smaller weed populations, each of our online planning algorithms - JUMP, SNAKE and R-SNAKE - lead to large improvement in path length, by up to 60\%, as compared to BCP and REACT. As the weed density increases beyond a threshold, it is more efficient to mow the entire pasture instead of visiting individual weed locations. This is also apparent from the figure, as the SNAKE and R-SNAKE path lengths start to increase beyond BCP path length. However, even for large populations, the path length for JUMP planner does not increase proportionally and instead converges to BCP path length. This is expected since JUMP paths start to mimic BCP paths by design, as the weed density increases. We also observe that spatial distribution of weeds on the pasture does not affect the path length or weed coverage.

Figure \ref{fig:mowed_vs_weeds} shows additional insights on the effects of weed density. The percentage of weeds mowed by the random-search based reactive planner (REACT) drops drastically as the weed density increases since the effect of repetitive area coverage due to random sampling becomes more pronounced. JUMP and SNAKE planners on the other hand, ensure $100 \%$ and $99.4 \%$ weed coverage, respectively. R-SNAKE planner leads to 97.5 \% coverage on average, even though the detection rate for R-SNAKE is $\sim$100 \%. This is a result of the design philosophy for the planner that trades off weed coverage in favor of shorter path length and also explains the observation that R-SNAKE consistently has shorter path length than SNAKE.

\begin{figure}
\centering
\subfloat[]{\includegraphics[width = 0.5\linewidth]{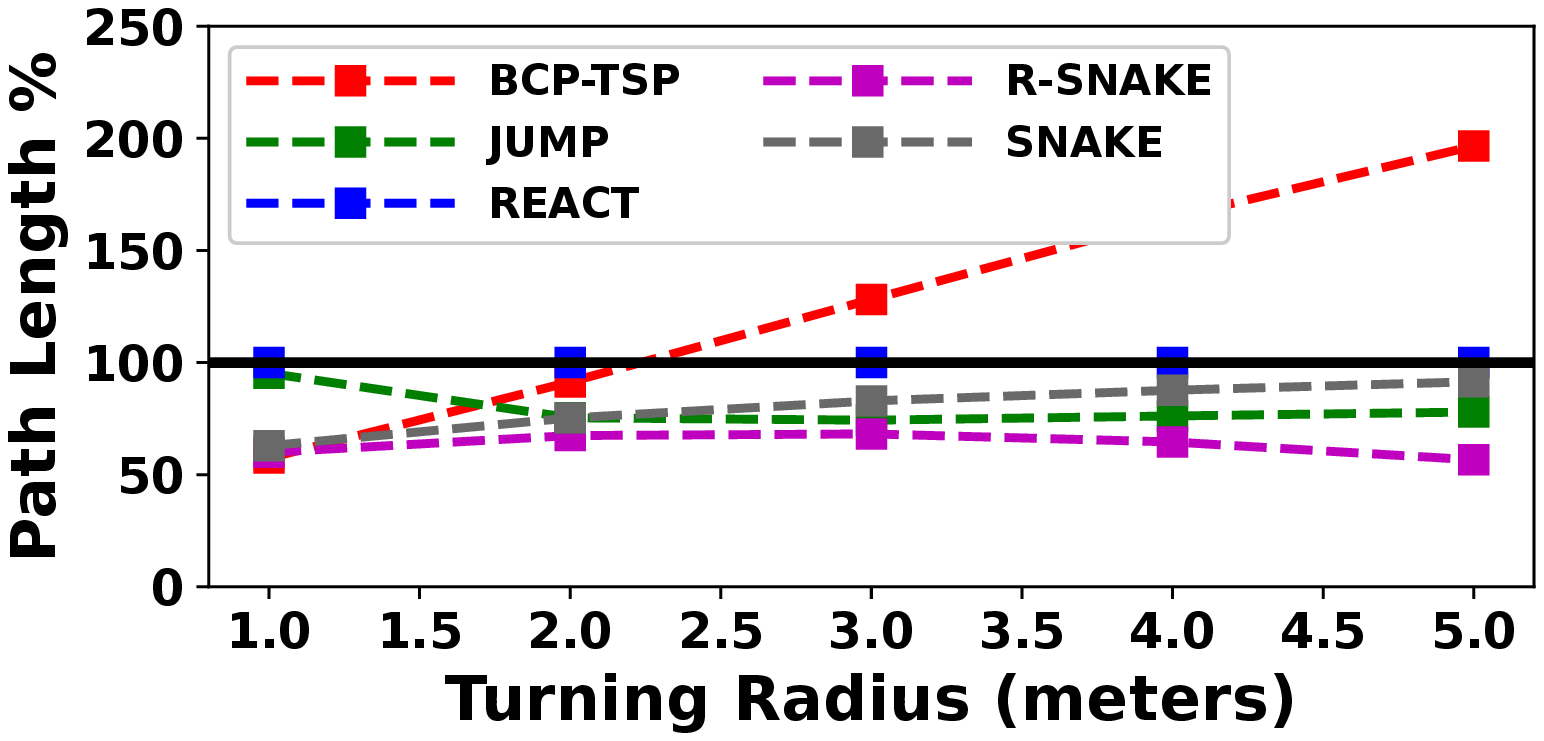}\label{fig:length_vs_radius}}
\subfloat[]{\includegraphics[width = 0.5\linewidth]{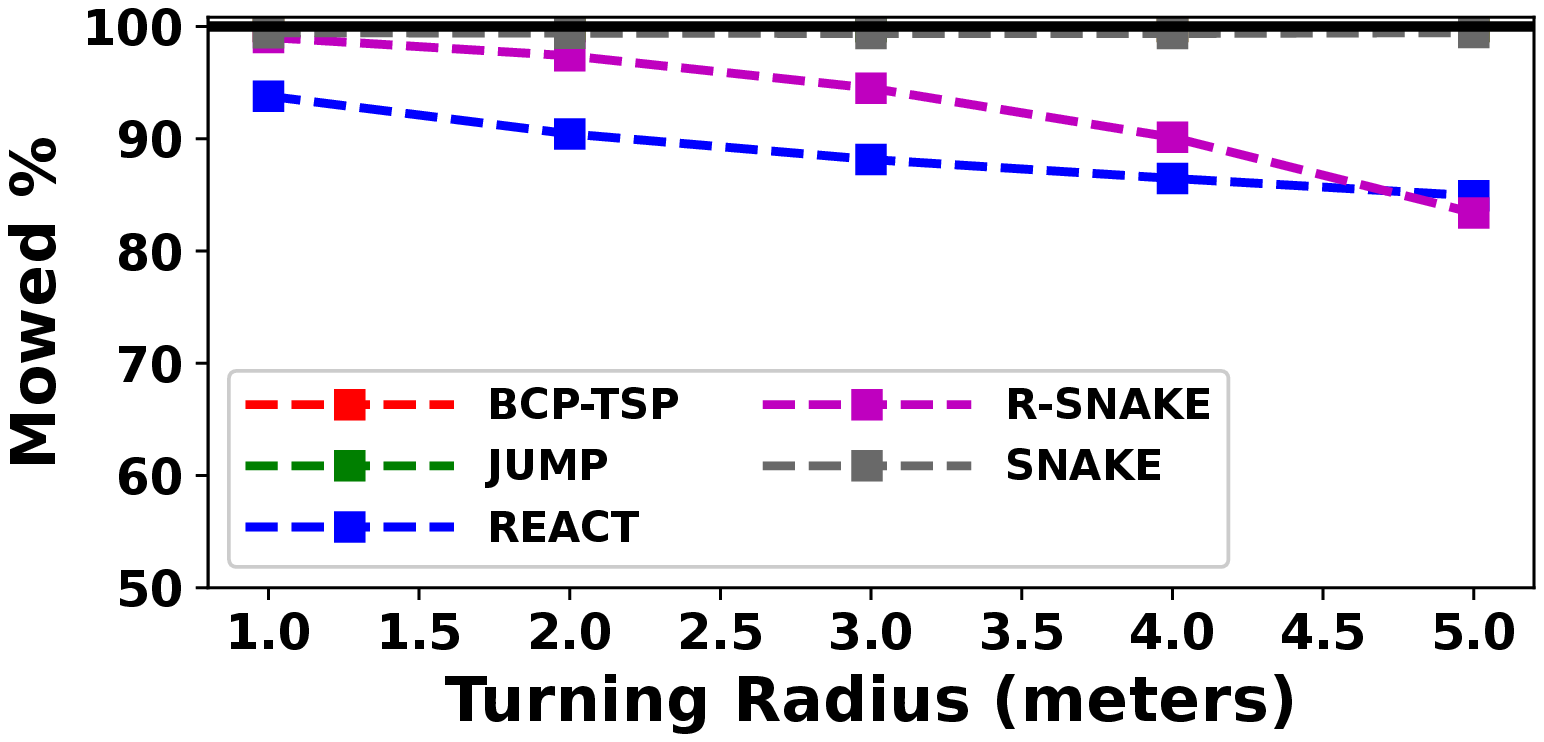}\label{fig:mowed_vs_radius}}
\caption{The plots show trends in (a) path length and (b) \% of weeds mowed, as a function of turning radius of the mower. The plots show data points averaged over all sizes of weed population. Fixed parameters: $\mathcal{S}_d = 12m, \mathcal{S}_w=12m$.}
\end{figure}

\subsubsection*{Effect of turning radius} We make two major observations from Figs. \ref{fig:length_vs_radius} and \ref{fig:mowed_vs_radius} showing the effects of turning radius on the performance of the planning algorithms. First, the weed coverage of R-SNAKE planner decreases with increase in turn radius. Large turn radius reduces the mower's agility and results in fewer subpaths on each pass. After a pass is completed, weeds that are towards the bottom of the pasture more than a fixed threshold below the mower are not considered anymore for mowing, thus increasing the number of unmowed weeds. Second, when the turning radius is very small, path length for the JUMP planner shows an abrupt increase. On investigation, we found that small turn radius makes it easier for the mower to compute jumps to mow nearby weeds. This increases the number of jumps on the mower's path, thereby increasing the path length.



\begin{figure*}
\centering
\subfloat[]{\includegraphics[width = 0.25\linewidth]{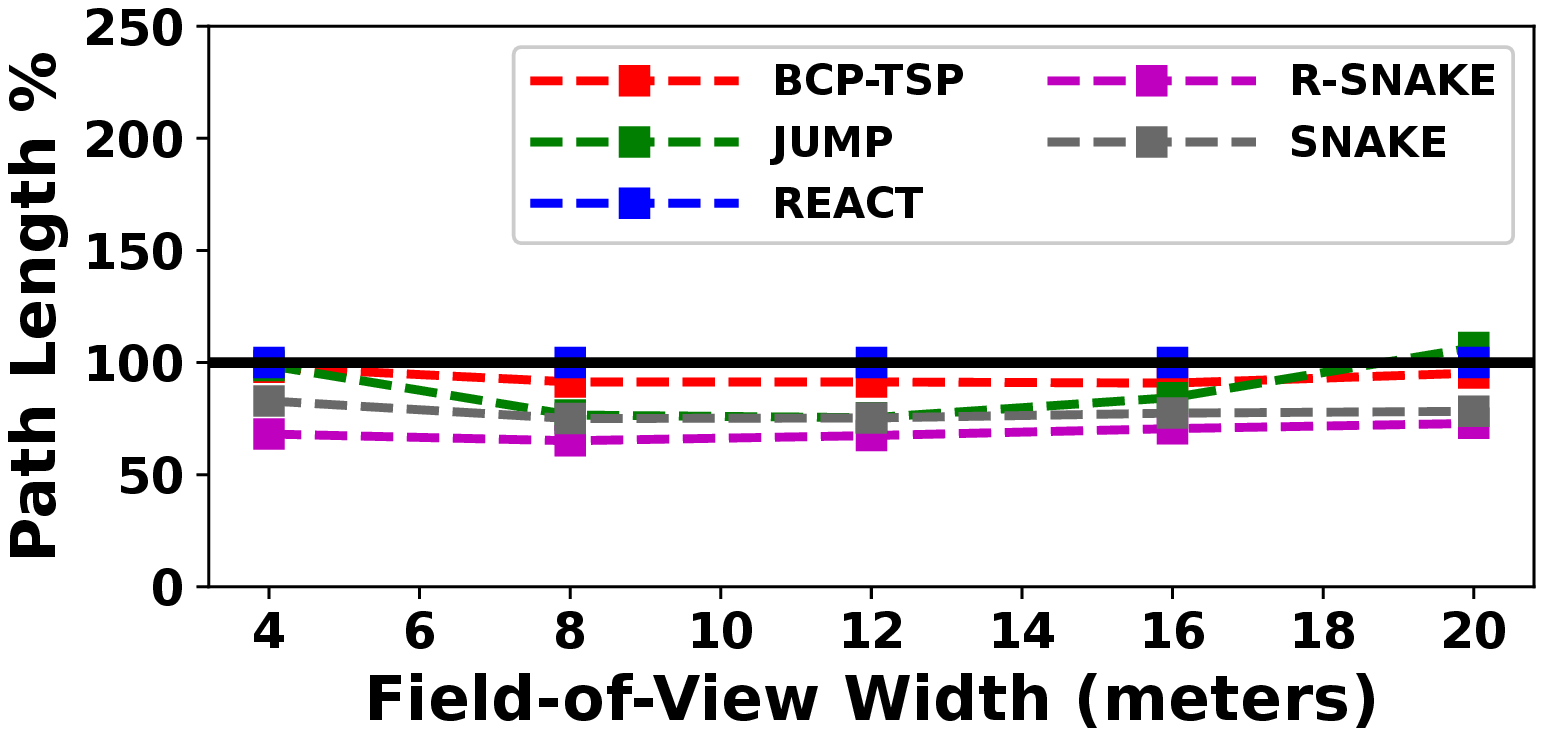}\label{fig:length_vs_width}}
\subfloat[]{\includegraphics[width = 0.25\linewidth]{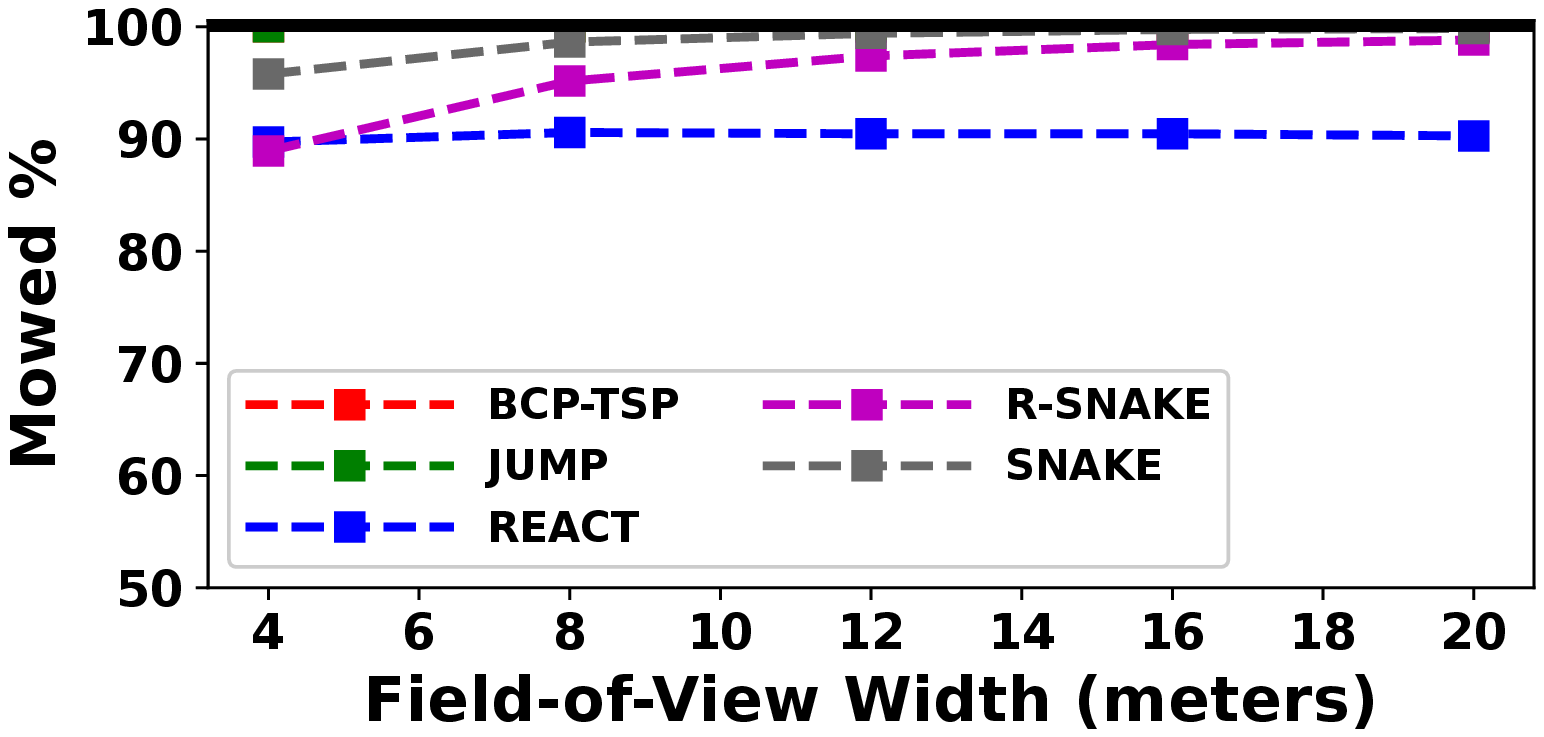}\label{fig:mowed_vs_width}}
\subfloat[]{\includegraphics[width = 0.25\linewidth]{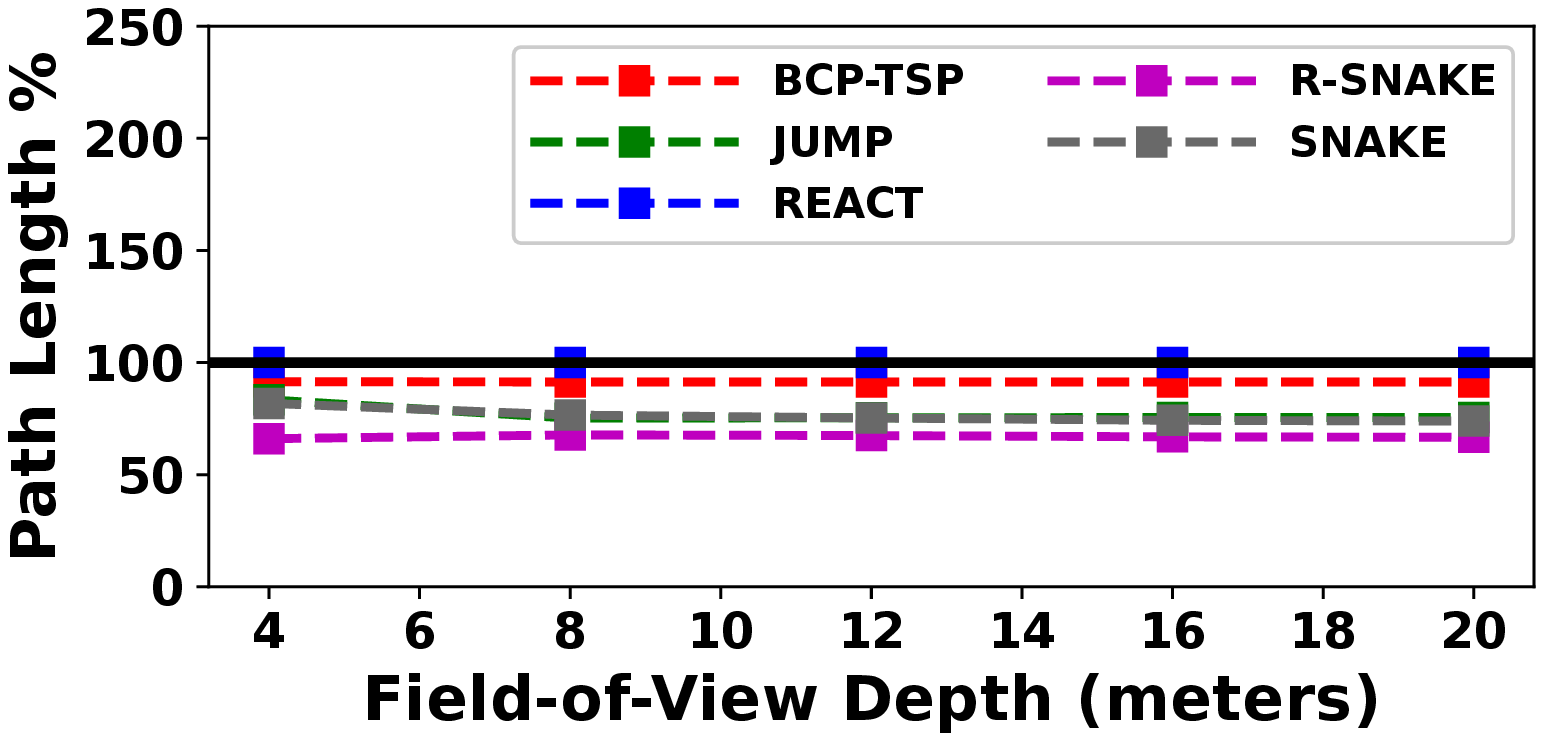}\label{fig:length_vs_depth}}
\subfloat[]{\includegraphics[width = 0.25\linewidth]{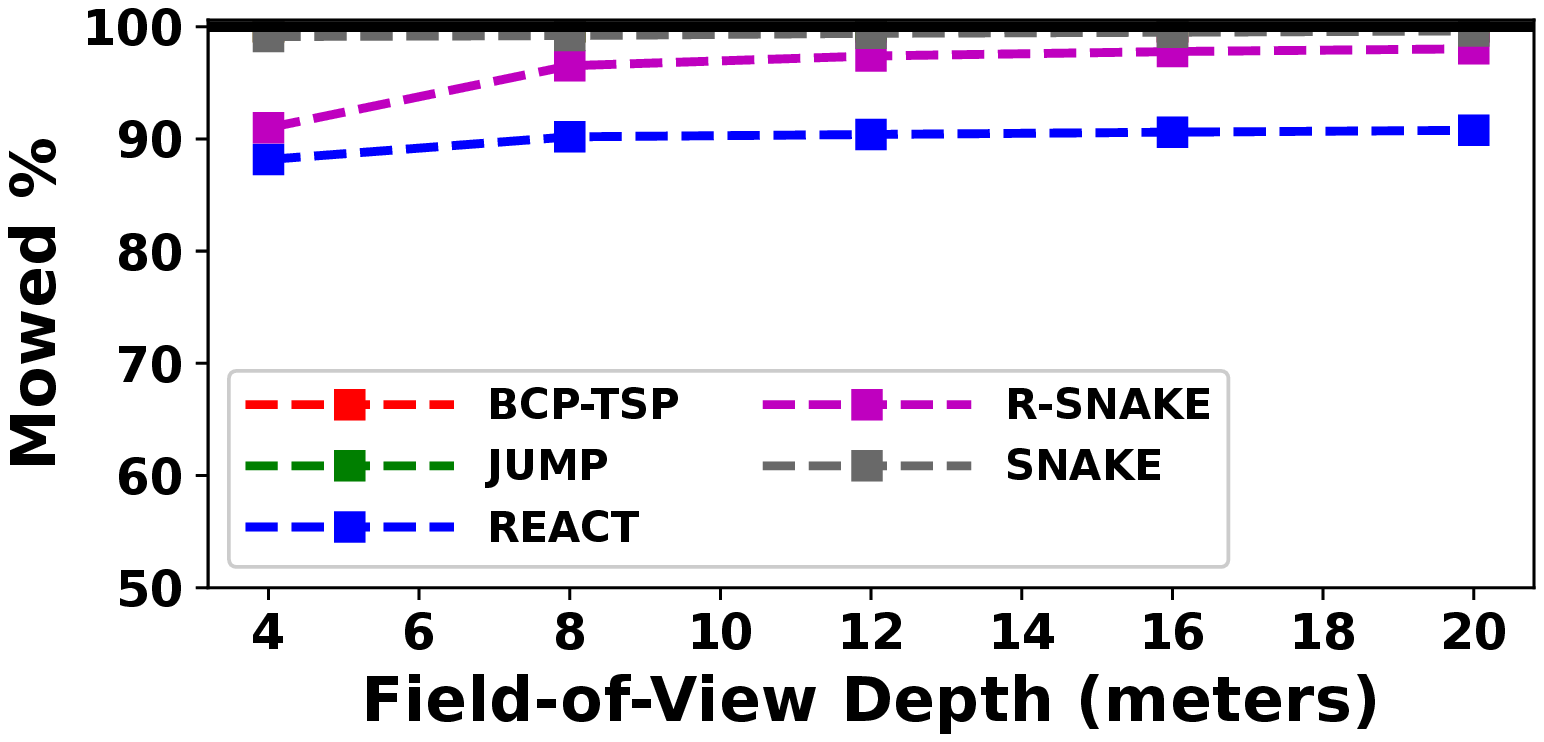}\label{fig:mowed_vs_depth}}
\caption{The plots show trends in path length, (a) and (c), and \% of weeds mowed, (b) and (d), as a function of change in FOV of the mower. (a) and (b) show variation with FOV width and (c) and (d) show variation with FOV depth. $\mathcal{R}$ was set to 2m. $\mathcal{S}_d$ was set to 12m for (a) and (b), and $ \mathcal{S}_w$ was set to 12m for (c) and (d).}
\end{figure*}

\subsubsection*{Effect of detection range}
Contrary to intuition, we found that the depth of the FOV does not have any noticeable effect on the path length. The width of the FOV, however, does impact path length for JUMP algorithm. We observe a parabolic trend in the path length of the JUMP planner when the width of the detection FOV, $\mathcal{S}_w$, is varied (Figs. \ref{fig:length_vs_width} and \ref{fig:mowed_vs_width}). On investigation, we found that $\mathcal{S}_w$ affects the detection of weeds that do not lie on the current pass and result in jumps on the mower's path. When the FOV is narrow, the path has negligible number of jumps and converges to a BCP path. When the FOV is wide, the mower's path comprises of excessive number of jumps resulting in an increase in the path length. In the case of R-SNAKE, small FOV caused by either of depth or width reduces the weed coverage.

\section{Field Experiments}\label{sec:exp_results}

\begin{figure}
\centering
\subfloat[Before Mowing]{\includegraphics[height = 1cm]{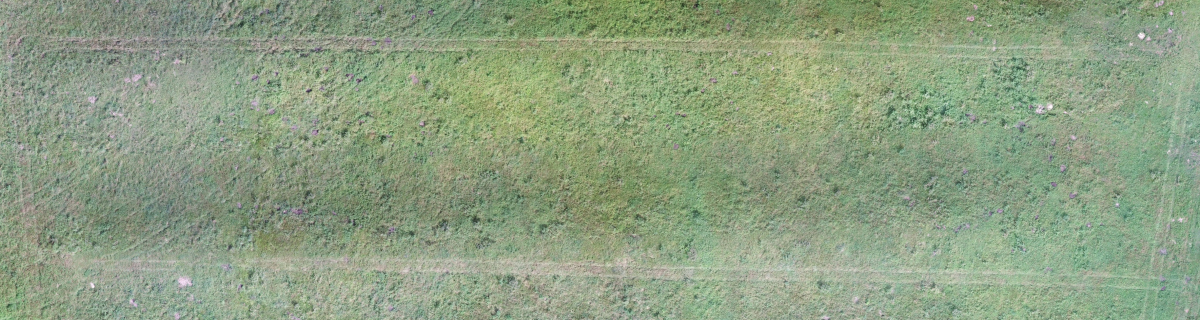}}\hspace{0.1mm}
\subfloat[After Mowing]{\includegraphics[height = 1cm]{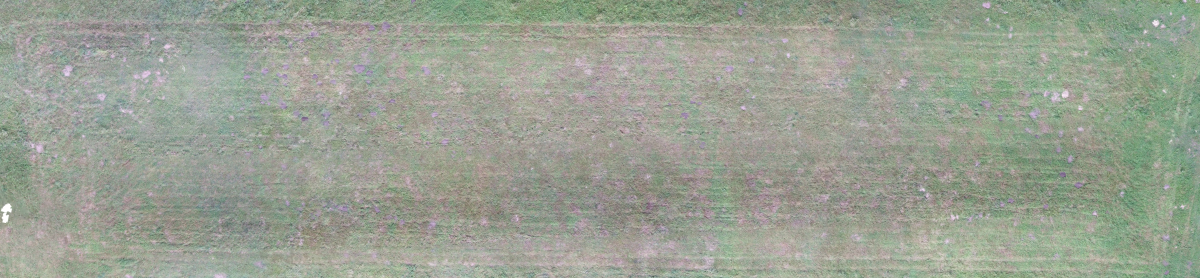}}
\caption{Before and after aerial views of a cow pasture of size $90 \times 20$ sq. meters mowed by the Cowbot with a boustrophedon coverage path. The pasture boundary in the before image is highlighted by a mowed perimeter.}
\label{fig:aerialBCP}
\end{figure}

Cowbot has been tested extensively on multiple farm fields with different types of terrain. It is able to mow large and dense weed populations and is able to negotiate rugged environments in cow pastures. It uses a navigation controller inspired by the Stanley controller \cite{stanley} for waypoint-navigation. The controller was tuned extensively in environments with varying topography including grass fields, steep hills, cow pastures with dense vegetation and rough terrain. The cut-quality of Cowbot and performance of the navigation controller were tested using BCP paths on cow pastures. Figure \ref{fig:aerialBCP} shows aerial views of a cow pasture located in Morris, Minnesota before and after a Cowbot BCP mowing experiment. To evaluate the performance of the online planning algorithms for Cowbot, field experiments were conducted on a farm field located near Morgan in Minnesota, United States during Cowbot's public demonstrations at the Minnesota Farm Fest 2021. The experiments were conducted in an area of size $36\times26$ sq. meters. 

\begin{figure}
\centering
\subfloat[JUMP]{\includegraphics[width = 0.33\linewidth]{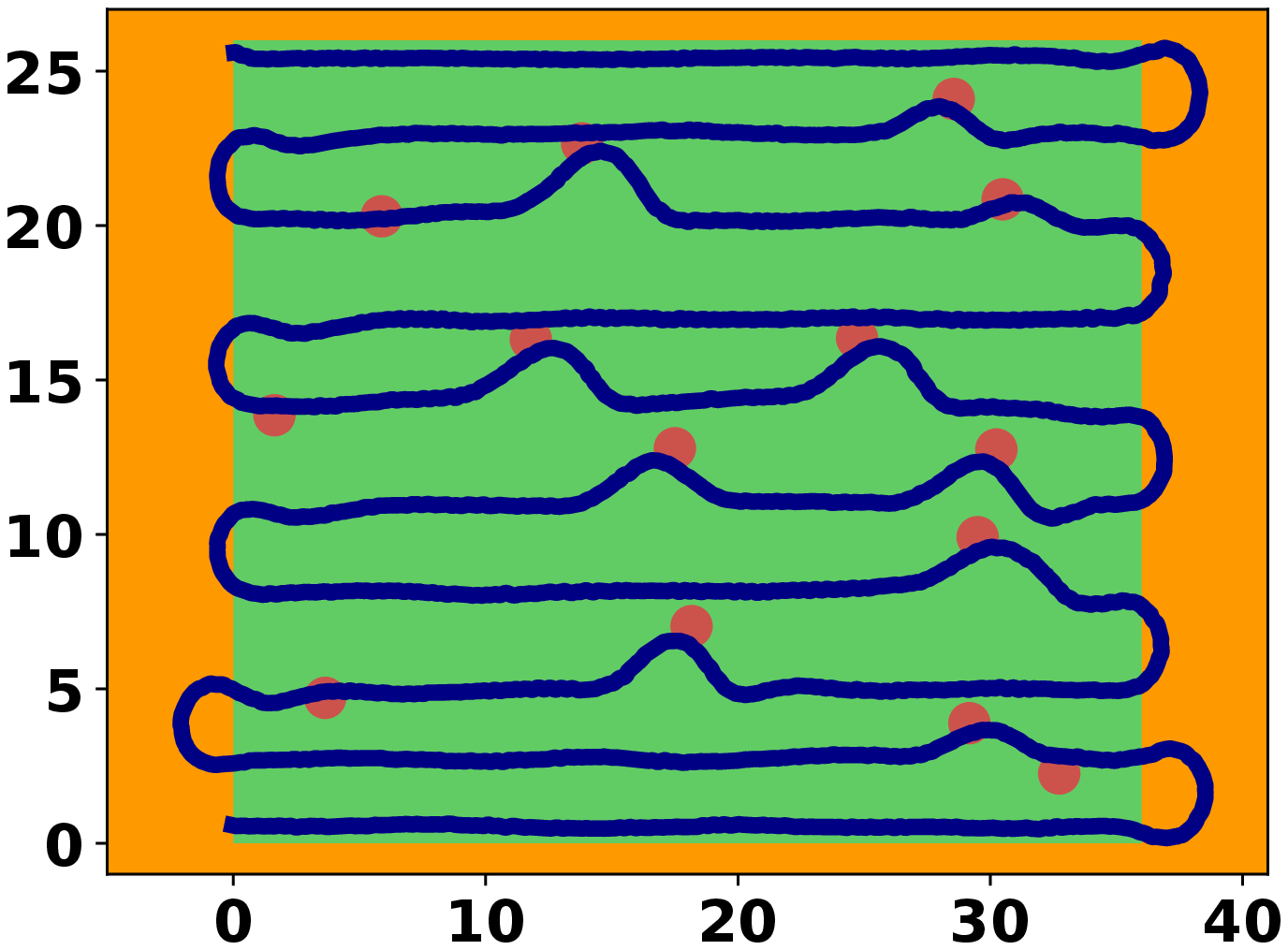}}
\subfloat[SNAKE]{\includegraphics[width = 0.33\linewidth]{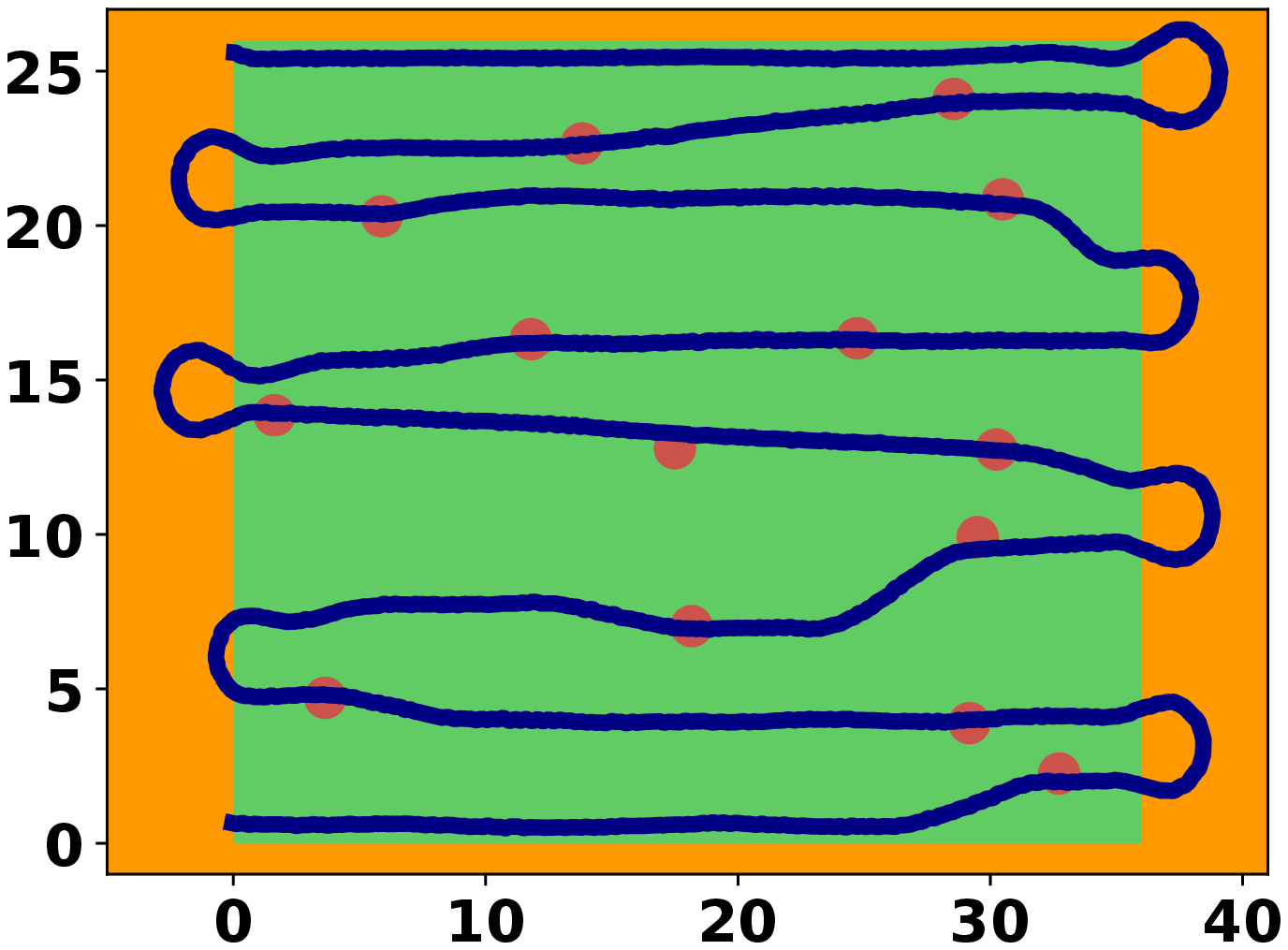}}
\subfloat[R-SNAKE]{\includegraphics[width = 0.33\linewidth]{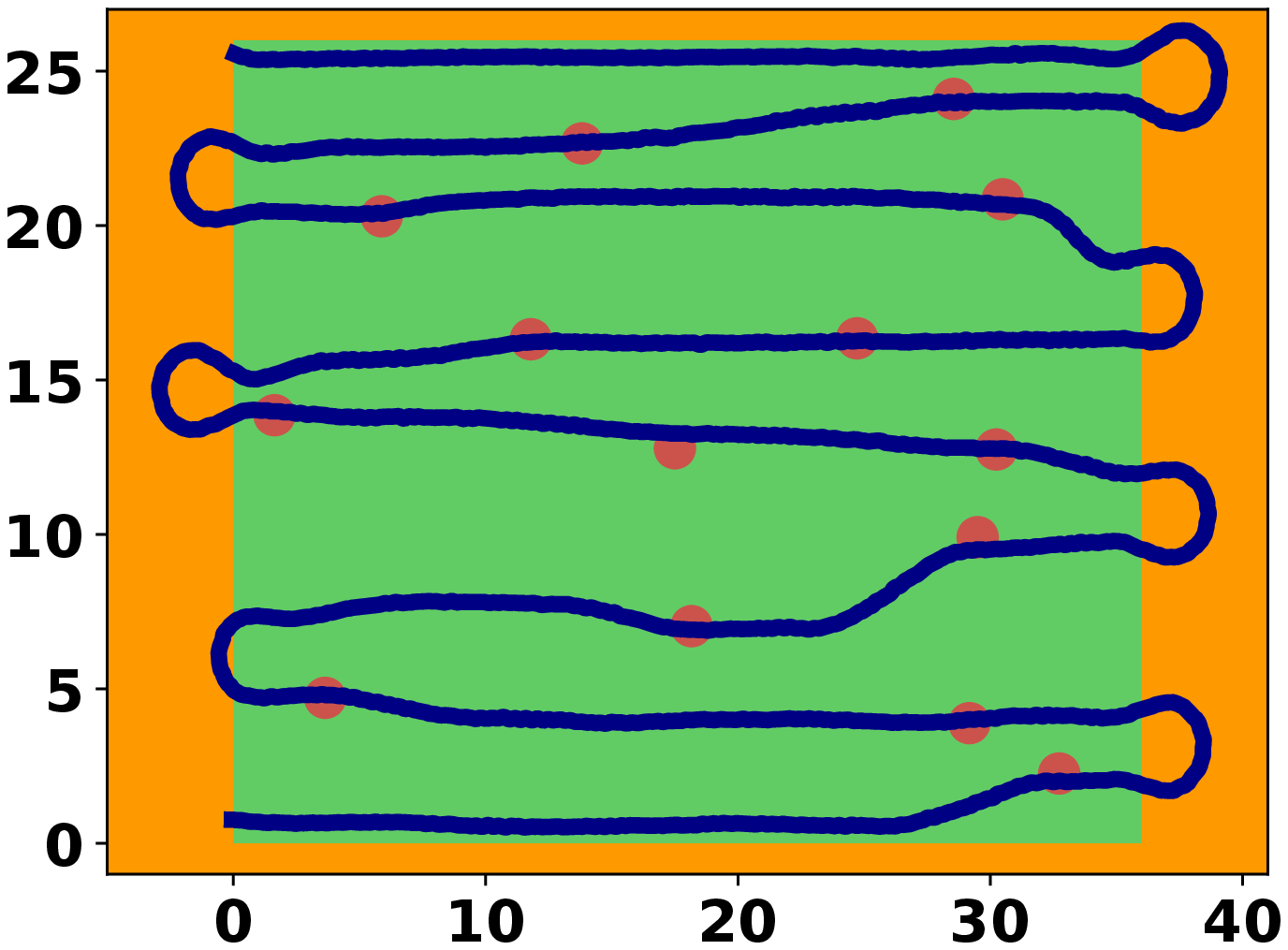}}
\caption{Visualization of the path followed by Cowbot in field experiments using different online planning algorithms. Green area marks the boundary of the pasture area and red disks show the location of weed markers on the field.}
\label{fig:expPathPlots}
\end{figure}

For the purpose of these experiments we assume the availability of a weed detection module with a detection FOV of size $(S_d,S_w)\equiv(12,12)$, similar in configuration to the mower model discussed in Section \ref{sec:sol_methods}. To visualize the experiments we geotagged 14 weed locations, sampled uniformly at random, on the field and marked them with flags. Cowbot's instantaneous FOV was computed based on its location and heading estimates. When a geotagged weed entered the FOV of the mower it was marked as detected and it's location information was made available to the planner. We conducted field experiments for JUMP, SNAKE and R-SNAKE planning algorithms. The BCP path length computed offline for the pasture is 897 m. Length of JUMP, SNAKE and R-SNAKE paths followed by the Cowbot on the field were 410 m (45.7\%), 338 m (37.6\%) and 338 m (37.6\%) respectively. The numbers in bracket show the relative path length to the BCP path. Each of the mower paths were able to mow all 14 weeds on the field. Figure \ref{fig:expPathPlots} traces the paths followed by the Cowbot for each of the three online planners. The proof-of-concept experiments show that Cowbot is suitable for use in organic cow pastures and our online planning algorithms are computationally efficient to deploy on real-time systems. Experiment video showing the Cowbot in action are included in the multimedia attachment.


\section{Conclusion and Future Directions}\label{sec:concFut}
Dairy pastures provide a unique environment amongst agricultural farms. They can be highly rugged and difficult to traverse. Lawn mowers and small research robots are not well suited for operating in these environments. We presented a novel autonomous weed mowing robot called Cowbot which provides a robust and stable platform to operate on rugged cow pastures. 

The density and distribution of weeds on a pasture can vary significantly. As a result, coverage paths do not always provide the most efficient solution. To address this online path planning problem we introduced the mower routing problem and developed online path planning algorithms, named JUMP and SNAKE (and R-SNAKE as a variant of SNAKE), to efficiently detect and mow weed on the pasture. We deployed our algorithms on the Cowbot and performed field validation experiments. Our online planners were shown to be computationally efficient and suitable for real-time path planning for mowing weeds. We also performed large scale simulated experiments to study the effect of various parameters on the performance of our methods. We discussed the trade-offs of our planning algorithms and the conditions that each of them is best suited for.

We are currently working on incorporating on board computer vision algorithms to detect weeds on the fly. In our future work we would also like to address energy efficiency, robustness and safety of Cowbot.


\bibliographystyle{IEEEtran}
\bibliography{reference}
\end{document}